\documentclass[10pt,journal,compsoc]{IEEEtran}

\usepackage{array,multirow}
\usepackage{booktabs} 
\usepackage{graphicx}
\usepackage{caption}
\usepackage{subfig}
\usepackage{amssymb}
\usepackage{threeparttable}
\usepackage{booktabs}
\usepackage[nocompress]{cite}
\usepackage{bm}
\usepackage{amsmath,amssymb,amsfonts}
\usepackage{algorithmic}
\usepackage{graphicx}
\usepackage{mathtools}
\usepackage{textcomp}
\usepackage{xcolor}
\usepackage{physics}
\usepackage[hyphens]{url}

\usepackage[capitalize]{cleveref}
\crefname{section}{Sec.}{Secs.}

\def\BibTeX{{\rm B\kern-.05em{\sc i\kern-.025em b}\kern-.08em
    T\kern-.1667em\lower.7ex\hbox{E}\kern-.125emX}}

\usepackage[acronyms,nonumberlist,nopostdot,nomain,nogroupskip]{glossaries}
\newacronym{cppomdp}{CP-POMDP}{Cyber-Physical POMDP}
\newacronym{decpomdp}{Dec-POMDP}{Decentralized POMDP}
\newacronym{cdma}{CDMA}{Code Division Multiple Access}
\newacronym{aoi}{AoI}{Age of Information}
\newacronym{aoii}{AoII}{Age of Incorrect Information}
\newacronym{voi}{VoI}{Value of Information}
\newacronym{uoi}{UoI}{Urgency of Information}
\newacronym{iot}{IoT}{Internet of Things}
\newacronym{marl}{MARL}{Multi-Agent Reinforcement Learning}
\newacronym{drl}{DRL}{Deep Reinforcement Learning}
\newacronym{ml}{ML}{Machine Learning}
\newacronym{rl}{RL}{Reinforcement Learning}
\newacronym{pomdp}{POMDP}{Partially Observable Markov Decision Process}
\newacronym{uav}{UAV}{Unmanned Aerial Vehicle}
\newacronym{dqn}{DQN}{Deep Q-Network}
\newacronym{dial}{DIAL}{Differentiable Inter-Agent Learning}
\newacronym[plural=MDPs,firstplural=Markov Decision Processes (MDPs)]{mdp}{MDP}{Markov Decision Process}
\newacronym{fov}{FoV}{Field of View}
\newacronym{cnn}{CNN}{Convolutional Neural Network}
\newacronym{nn}{NN}{Neural Network}
\newacronym{ddql}{DDQL}{Distributed Deep Q-Learning}
\newacronym{pdf}{PDF}{Probability Density Function}
\newacronym{ndpomdp}{ND-POMDP}{Networked Distributed POMDP}
\newacronym{radam}{RAdam}{Rectified Adam}
\newacronym{cdf}{CDF}{Cumulative Distribution Function}
\newacronym{mpc}{MPC}{Model Predictive Control}
\newacronym{rv}{rv}{Random Variable}
\newacronym{qoe}{QoE}{Quality of Experience}
\newacronym{qos}{QoS}{Quality of Service}
\newacronym{fec}{FEC}{Forward Error Correction}
\newacronym{relu}{ReLU}{Rectified Linear Unit}
\newacronym{cps}{CPS}{Cyber-Physical System}

\newacronym{nc}{NC}{Networked Control}
\newacronym{jcc}{JCC}{Joint Communication and Control}

\newacronym{rc}{RC}{ \textit{Random Communication}}
\newacronym{sc}{SC}{ \textit{Sequential Communication}}
\newacronym{cc}{CC}{ \textit{Closest Communication}}
\newacronym{oc}{OC}{ \textit{Oracle Communication}}

\newacronym{cjcc}{CJCC}{Centralized JCC}
\newacronym{djcc}{DJCC}{Distributed JCC}

\newacronym{adam}{Adam}{Adaptive moment estimator}
\newacronym{auv}{AUV}{Autonomous Underwater Vehicle}

\usepackage{tikz}
\usepackage{pgfplots}
\usepackage{tikzscale}
\usepackage{tikz-qtree}
\usepackage{tabularx}
\pgfplotsset{compat=newest}
\pgfplotsset{plot coordinates/math parser=false}
\pgfplotsset{every axis/.append style={
                    label style={font=\scriptsize},
                    tick label style={font=\scriptsize},
                    legend style={font=\scriptsize}
                    }}
\usetikzlibrary{plotmarks,shapes,patterns,decorations.pathreplacing,backgrounds,calc,arrows,arrows.meta,spy,matrix,shadows,trees,positioning,fit}
\usepgfplotslibrary{patchplots,groupplots}

\tikzstyle{startstop} = [rectangle, rounded corners, minimum width=2cm, minimum height=0.5cm,text centered, draw=black]
\tikzstyle{io} = [trapezium, trapezium left angle=70, trapezium right angle=110, minimum width=3cm, minimum height=1cm, text centered, draw=black]
\tikzstyle{process} = [rectangle, minimum width=2cm, minimum height=0.5cm, text centered, draw=black, alignb=center]
\tikzstyle{decision} = [ellipse, minimum width=2cm, minimum height=1cm, text centered, draw=black]
\tikzstyle{arrow} = [thick,<->,>=stealth]
\tikzstyle{line} = [thick,>=stealth]
\tikzstyle{darrow} = [thick,<->,>=stealth,dashed]
\tikzstyle{sarrow} = [thick,->,>=stealth]
\tikzstyle{larrow} = [line width=0.1mm,dashdotted,->,>=stealth]

\pgfkeys{/pgf/number format/.cd,1000 sep={\,}} 

\makeatletter
\def\grd@save@target#1{%
  \def\grd@target{#1}}
\def\grd@save@start#1{%
  \def\grd@start{#1}}
\tikzset{
  grid with coordinates/.style={
    to path={%
      \pgfextra{%
        \edef\grd@@target{(\tikztotarget)}%
        \tikz@scan@one@point\grd@save@target\grd@@target\relax
        \edef\grd@@start{(\tikztostart)}%
        \tikz@scan@one@point\grd@save@start\grd@@start\relax
        \draw[minor help lines] (\tikztostart) grid (\tikztotarget);
        \draw[major help lines] (\tikztostart) grid (\tikztotarget);
        \grd@start
        \pgfmathsetmacro{\grd@xa}{\the\pgf@x/1cm}
        \pgfmathsetmacro{\grd@ya}{\the\pgf@y/1cm}
        \grd@target
        \pgfmathsetmacro{\grd@xb}{\the\pgf@x/1cm}
        \pgfmathsetmacro{\grd@yb}{\the\pgf@y/1cm}
        \pgfmathsetmacro{\grd@xc}{\grd@xa + \pgfkeysvalueof{/tikz/grid with coordinates/major step x}}
        \pgfmathsetmacro{\grd@yc}{\grd@ya + \pgfkeysvalueof{/tikz/grid with coordinates/major step y}}
        \foreach \x in {\grd@xa,\grd@xc,...,\grd@xb}
        \node[anchor=north] at (\x,\grd@ya) {\pgfmathprintnumber{\x}};
        \foreach \y in {\grd@ya,\grd@yc,...,\grd@yb}
        \node[anchor=east] at (\grd@xa,\y) {\pgfmathprintnumber{\y}};
      }
    }
  },
  minor help lines/.style={
    help lines,
    gray,
    line cap =round,
    xstep=\pgfkeysvalueof{/tikz/grid with coordinates/minor step x},
    ystep=\pgfkeysvalueof{/tikz/grid with coordinates/minor step y}
  },
  major help lines/.style={
    help lines,
    line cap =round,
    line width=\pgfkeysvalueof{/tikz/grid with coordinates/major line width},
    xstep=\pgfkeysvalueof{/tikz/grid with coordinates/major step x},
    ystep=\pgfkeysvalueof{/tikz/grid with coordinates/major step y}
  },
  grid with coordinates/.cd,
  minor step x/.initial=.5,
  minor step y/.initial=.2,
  major step x/.initial=1,
  major step y/.initial=1,
  major line width/.initial=1pt,
}
\makeatother

\newlength\fheight
\newlength\fwidth

\makeatletter
\def\endthebibliography{%
	\def\@noitemerr{\@latex@warning{Empty `thebibliography' environment}}%
	\endlist
}
\makeatother

\definecolor{color0}{HTML}{FFD700}
\definecolor{color1}{HTML}{FFB14E}
\definecolor{color2}{HTML}{FA8775}
\definecolor{color3}{HTML}{EA5F94}
\definecolor{color4}{HTML}{CD34B5}
\definecolor{color5}{HTML}{9D02D7}
\definecolor{color6}{HTML}{0000FF}
\definecolor{violet}{rgb}{0.6,0,0.6}%
\definecolor{orange_D}{rgb}{1,0.3,0}%
\definecolor{cyan}{rgb}{0,0.67,0.64}%
\definecolor{red}{rgb}{0.9,0,0}%
\definecolor{green_D}{rgb}{0,0.5,0}%
\definecolor{yellow}{rgb}{1,0.8,0}

\def \fwidth{0.6\columnwidth}
\def \fheight {0.2\columnwidth}

\begin{document}

\title{Multi-Agent Reinforcement Learning for Pragmatic Communication and Control}


\author{Federico Mason,~\IEEEmembership{Student Member,~IEEE,} Federico Chiariotti,~\IEEEmembership{Member,~IEEE,}  \\
Andrea Zanella,~\IEEEmembership{Senior Member, IEEE,} and Petar Popovski,~\IEEEmembership{Fellow,~IEEE}
\IEEEcompsocitemizethanks{\IEEEcompsocthanksitem{Federico Mason (email: federico.mason@unibo.it) is also with the Department of Biomedical and Neuromotor Science, University of Bologna, Italy. He is also affiliated with the Department of Information Engineering, University of Padova, Padua, Italy, as are Federico Chiariotti (corresponding author, email: chiariot@dei.unipd.it) and Andrea Zanella (email: zanella@dei.unipd.it). Petar Popovski (email: petarp@es.aau.dk) are with the Department of Electronic Systems, Aalborg University, Aalborg, Denmark, with which Federico Chiariotti is also affiliated.} \IEEEcompsocthanksitem{This work was partly financed by the Velux Foundation under the WATER Villum Investigator grant and by the European Union under the Italian National Recovery and Resilience Plan (NRRP) of NextGenerationEU, under the REDIAL Young Researchers grant and the partnership on “Telecommunications of the Future” (PE0000001 - program
“RESTART”).}}}
\markboth{IEEE Transactions on Mobile Computing,~Vol.~xx, No.~x, Dec. 2022}%
{Mason \MakeLowercase{\textit{et al.}}: Joint Goal-Oriented Communication and Control in Cooperative Multi-Agent Problems}

\IEEEtitleabstractindextext{%
\begin{abstract}
The automation of factories and manufacturing processes has been accelerating over the past few years, boosted by the Industry 4.0 paradigm, including diverse scenarios with mobile, flexible agents. Efficient coordination between mobile robots requires reliable wireless transmission in highly dynamic environments, often with strict timing requirements. Goal-oriented communication is a possible solution for this problem: communication decisions should be optimized for the target control task, providing the information that is most relevant to decide which action to take. From the control perspective, networked control design takes the communication impairments into account in its optmization of physical actions. In this work, we propose a joint design that combines goal-oriented communication and networked control into a single optimization model, an extension of a multiagent \gls{pomdp} which we call \acrfull{cppomdp}. The model is flexible enough to represent several swarm and cooperative scenarios, and we illustrate its potential with two simple reference scenarios with a single agent and a set of supporting sensors. Joint training of the communication and control systems can significantly improve the overall performance, particularly if communication is severely constrained, and can even lead to implicit coordination of communication actions.
\end{abstract}

}

\IEEEdisplaynontitleabstractindextext

\maketitle

\glsresetall

\IEEEraisesectionheading{\section{Introduction}}

Over the last decade, the use of \gls{iot} technology for industrial and manufacturing applications has exploded~\cite{cheng2018industrial}. The digitization and automation of industrial processes is a key component of the Industry 4.0 paradigm~\cite{wollschlaeger2017future}, and the coordination of mobile robots requires the use of wireless communications.
The optimization of these systems has been approached from two opposite directions: from the control side, we have seen the development of networked control systems~\cite{zhang2017analysis} which account for the \gls{qos} of the communication system in the controller design, using metrics such as the latency distribution or the packet error rate. From the communication perspective, growing attention has been dedicated to new performance indicators that can capture the communication needs of control applications, guiding the optimization of scheduling and resource allocation~\cite{klugel2019aoi}. A representative indicator is \gls{aoi}~\cite{kaul2012real}, which measures the freshness of the information available to the controller.

Another significant development in the field has been the explosion of \gls{rl} techniques, which began in 2015 with DeepMind's seminal paper on \glspl{dqn}~\cite{mnih2015human}. The union of deep learning with reinforcement techniques has triggered a revolution in robotics, as \glspl{dqn} are more flexible and adaptable than human-designed techniques in a variety of complex situations. Naturally, the \gls{dqn} is optimized for its training scenario, and will effectively act as a networked control system, countering the impairments caused by the wireless connection and maximizing performance, if it is trained in that environment. However, the inherent complexity of these learning agents complicates the design of the communication system. Since learning agents operate as black boxes, it is difficult to know which piece of information will be crucial and which will be almost irrelevant~\cite{ayan2019age}, reducing the effectiveness of \gls{aoi}-based approaches.

The quest for sending relevant and meaningful information has lead to an increased interest in semantic communications. These approaches are commonly inspired by Weaver's introduction to the expanded version of Shannon's seminal paper~\cite{shannon1948mathematical}. In this text, he classifies the traditional communication problem that targets a set of \gls{qos} measures as a \emph{technical} problem and goes on to define two types of problems beyond it, which are the \emph{semantic} and \emph{effectiveness} problems, respectively. If we are to draw a full parallel to linguistics, then the effectiveness corresponds to \emph{pragmatics}, where  the ultimate goal of communication is to affect the behavior of intelligent agents, taking into account shared context and implicit clues. While effectiveness has been discussed for decades~\cite{ritchie1986shannon}, the complexity of defining and solving it has limited the possibility of practical solutions until recently, when the massive improvement of machine learning techniques has allowed researchers to tackle it. However, this research domain is still in its infancy, and learning to control a system and communicate efficiently at the same time is an open problem for non-trivial scenarios~\cite{tung2021effective}.

In this paper, we propose a general model of a \gls{cps} that integrates communication and control, allowing for their joint optimization using \gls{marl}~\cite{lazaridou2020emergent}. More specifically, we consider a dynamic environment in which multiple agents interact. There are two agent types:  \emph{robots} that can perform actions and affect the physical environment; and \emph{sensors}, whcich communicate their observations to the robots through resource-constrained communication channels. This maps directly to the effectiveness problem, as the ultimate goal is to optimize the control of the robots by exploiting the data from the sensors to maintain an accurate picture of the state of the system. Our model follows the \gls{pomdp} theory, and the joint communication and control aspects, as well as the interaction between distributed agents, led us to name it \gls{cppomdp}. Some existing models, such as \glspl{ndpomdp}, include communications between agents, but do not optimize them directly.

Our model captures the direct and indirect aspects of communication, as sharing information can affect both the physical environment and the observations available to each agent about it.
More importantly, control and communication decisions are both captured by the model, allowing the joint training of agents that perform one or both types of actions, i.e., interact with the physical environment and communicate local information to other nodes.
To the best of our knowledge, this is the first model for \gls{marl} that considers the dual nature of the problem, optimizing communications and control together. As it will be seen, the solution follows the principle that coordination is implicit communication, that is, there is a blurred border between acts of communication and acts of control.
The objective of this paper is hence to effectively bridge the gap between approaches from both communication and control perspective, joining the two into a single comprehensive \gls{marl} model, which is expressive yet lightweight.

The benefits of solving the effectiveness problem are more pronounced when the  communication is constrained and the data that gets actually transmitted should be carefully selected. This is typical in underwater scenarios with acoustic communications~\cite{zia2021state}, which are used in this paper to showcase our \gls{marl} solution. In both scenarios, an \gls{auv} needs to reach a surface vessel as fast as possible after performing a mission, and static sensor buoys can expand its field of view by communicating their observations. As the acoustic medium can only support very low bit rates, the communication from the buoys is necessarily limited. The mission in the first scenario involves avoiding obstacles, such as  underwater installations or natural rock formations, even sea mines. The second mission is a data muling scenario~\cite{fu2022reinforcement} in which the \gls{auv} needs to recover data from floating sensors that might move with the current. We then compare some heuristic optimization schemes for the communication system (such as scheduling the closest buoy at any time) with our joint optimization. The results show that joint optimization can significantly improve the performance of the \gls{auv} both when the communication decisions are made by a centralized controller with full knowledge of the map and when they are made in a distributed fashion by the individual buoys. In the latter, the buoys can learn to provide the most valuable information to the \gls{auv} and avoid packet collisions. The implications of the proposed method are not limited to this simple underwater models and we provide some examples of larger and more complex \glspl{cppomdp}.

The rest of this paper is organized as follows: first, we present the state of the art on \gls{aoi} optimization and networked control systems in Sec.~\ref{sec:related}, then define the \gls{cppomdp} model in Sec.~\ref{sec:cppomdp}. We then focus on the case with a single robot, aided by a number of sensors: the system model is rigorously defined as an instance of a \gls{cppomdp} in Sec.~\ref{sec:dp}. The two underwater tasks and the simulation settings and results are presented in Sec.~\ref{sec:analysis}, and finally, Sec.~\ref{sec:concl} concludes the paper and lists some possible avenues of future work on the subject.

\section{Related Work}\label{sec:related}

The growing use of wirelessly connected devices in control and monitoring scenarios with significant timing and resource constraints, such as industrial automation, coordination between autonomous cars, and drone swarm management, has led to the definition of new metrics, such as the~\gls{aoi}~\cite{kosta2017age}. Unlike packet latency, which measures the time from a packet's generation to its delivery to the intended receiver, ignoring the wider communication process (i.e., past and future transmission), the \gls{aoi} represents the freshness of the information available to the receiver, accounting for both the network latency and the rate at which the source generates data~\cite{kaul2012real}. \gls{aoi} has become widespread due to its intuitive relation with control processes: the freshness of the information when a control agent makes a decision represents the time delay between the environment estimated by the remote controller and the moment when the action is performed. Since its first introduction, there has been a plethora of works computing \gls{aoi} in different scenarios and optimizing communication systems to minimize the average or the tail of \gls{aoi} distribution~\cite{yates2020agesurvey}.

However, \gls{aoi} does not necessarily convey enough information to optimize intelligent networked control systems, as the underlying process that the transmitter is monitoring might remain stable over long periods of time, rendering the frequent updates as wasteful. Conversely, the process might sharply change, making the new information relevant even if an update has just been transmitted. In Markovian scenarios, in which the process can take discrete values, this issue is solved by using the \gls{aoii}~\cite{maatouk2020aoii}, which extends the notion of age to consider only updates that carry new information: as long as the process remains in the same state, new updates are not valuable. This concept can also be extended to continuous-valued and non-Markovian processes, leading to the more general concept of \gls{voi}~\cite{ayan2019age}: this metric is essentially an error function that depends on the difference between the estimated state at the receiver and the actual state of the process, which can integrate complex processes and predictive estimators~\cite{mason2020adaptive}.

Age and packet loss are not the only issues in networked control systems, as capacity can often become the most important constraint.
This is particularly important if sensors transmit video or other feature-rich data, often used in new vision-based control applications and can require a significant data rate.
In this case, the communication system needs to choose which pieces of information to transmit, and at which compression level~\cite{chiariotti2021quic}, in order to maximize the system performance while avoiding channel congestion.

However, even \gls{voi} might not be the optimal choice for networked control systems, as errors in the estimation of the system state may or may not translate to lower control performance. If the controller is highly sensitive to some error types, while relatively insensitive to others, defining the error function becomes complicated, and it is better to just use control performance as a proxy. This leads to the \gls{uoi} metric~\cite{zheng2020urgency}, which directly considers how much an update would affect a (known) controller. However, these metrics assume the controller to be fixed, while the communication system is optimized to accommodate it.

The work on networked control systems and \gls{marl} goes in the opposite direction, taking a fixed communication strategy and optimizing the controller (or controllers) in the system to deal with it. The most common method to deal with network impairments is to model a limited number of \gls{qos} metrics for the communication channel and design a controller that can perform well within a given operating region. Delay is the most common impairment that these systems consider, often assuming that packet delay follows a certain distribution and designing the controller to handle it~\cite{zhang2005new}. The difference between the delays in the two directions can also be taken into account, as in~\cite{lam2007stability}. Another important parameter in the design of networked control systems is the packet loss, as the erasure of an update can significantly affect the controller which relies on it to make its decisions~\cite{yue2005network}. However, errors can be dealt with by using reliable communication protocols, which retransmit lost packets, or by using \gls{fec}, trading additional latency for communication reliability~\cite{liu2015networked}. Naturally, the design of the controller is affected by both the features of the connection itself and the protocol~\cite{chiuso2014lqg}.

Finally, controllers need to deal with compression, often expressed as quantization~\cite{fridman2009control} or partial information, which can impair both inputs and outputs of the controller~\cite{zhang2017observer}.
Learning systems have recently been used to deal with random delays and errors~\cite{zhang2019iterative}, as their generalization capabilities allow them to better deal with unforeseen situations and compensate for the network impairments, particularly in non-linear systems~\cite{wu2017adaptive}.
Learning systems are also a viable option in case of compression of feedback or control signals~\cite{xu2015finite}; however, to the best of our knowledge, no unified framework for learning in networked control systems has been proposed.

The problem of communication and control has also been approached in \gls{marl}, either for the optimization of the communication system or for dealing with imperfect and delayed inputs. One of the first works to do so, by Foerster \emph{et al.}~\cite{foerster2016learning}, used simple problems that required coordination between agents, showing emergent communication capabilities. Other works use explicit or implicit communication costs so that agents can choose whether to coordinate, and either get a penalty or skip an action, or act independently~\cite{zhang2013coordinating}.

Two recent works are the most similar to our own: in the first~\cite{kim2019learning}, multiple agents learn the parameters of a simple scheduling algorithm while performing a reinforcement task in real time.
In that case, however, the use of a simple scheduling algorithm limits the applicability of the solution, whose maximum performance is unavoidably constrained by the communication policy.
Instead, our work uses the \gls{decpomdp} framework to model communication and control entirely within the \gls{marl} problem, making it possible to fully explore the system environment and identify new and more efficient solutions.
The second one~\cite{tung2021effective} considers a scenario similar to our own, with one sensor and one robot, in which the task involves learning the channel statistics and designing a shared codebook with \gls{rl} instead of learning which information to transmit. In some ways, this work is complementary to ours and could be combined with it to form a more complete communication and control system, as it focuses on encoding the available information instead of scheduling which part of it is more important in a distributed setting. For a more thorough review of the cooperative \gls{marl} literature, we refer the reader to~\cite{oroojlooyjadid2019review}.

\section{Joint Communication and Control}

\label{sec:cppomdp}

The problem addressed in this paper can be seen as a special case of the more general class of \glspl{decpomdp}.
While \glspl{decpomdp} considers communication implicitly, our model gives an explicit structure to the communication process among the system agents.
We name our model \gls{cppomdp}, as agents can both act in a physical environment and exchange information with other agents, coordinating to achieve a common goal.

\subsection{Reinforcement Learning}

In \glspl{mdp} and their extensions, time is discretized into slots so that, at each slot $k$, the environment is fully described by a state $s_k \in \mathcal{S}$.
In particular, an agent periodically observes the environment state and takes actions accordingly, affecting the system state evolution. 
Hence, at each slot $k$, the agent performs a new action $a(k) \in \mathcal{A}$, and receives a reward $r(k) \in \mathbb{R}$, depending on both $s(k)$ and $a(k)$. 
The agent behavior follows a policy $\pi: \mathcal{S} \times \mathcal{A} \rightarrow [0,1]$, where $\pi(s,a)$ is the probability to take action $a \in \mathcal{A}$ given the state $s \in \mathcal{S}$.
If the environment is only partially observable, the agent perceives the state through a function $\mathcal{S} \rightarrow \mathcal{O}$, associating each state $s$ with an observation $o$. The domain of the policy function is then given by the observation space $\mathcal{O}$. The goal of any \gls{rl} algorithm is to find a policy $\pi^*$ that maximizes the agent \emph{discounted return}, defined as
\begin{equation}
    G(k) = \sum_{\nu=0} \lambda^{\nu} r(k + \nu),
\end{equation}
where $\lambda \in [0,1)$ is an exponential \emph{discount factor}. 
To achieve this goal, an \gls{rl} algorithm needs to estimate the value of each possible observation-action pair $(o, a) \in \mathcal{O} \times \mathcal{A}$, determining the best action to take for each observation.
In other words, given a policy $\pi$, the algorithm has to approximate the function
\begin{equation}
    Q_{\pi}(o,a) = \mathbb{E}[G(k)|o(0)=0,a(0)=a,\pi],
\end{equation}
which is the expectation of the return obtained by choosing action $a$ when receiving observation $o$ and following the policy $\pi$ in the future. The \emph{Bellman optimality equation} ensures that the optimal policy $\pi^*$ provides the maximum value of $Q_{\pi}(o,a)$ in each possible condition, i.e., 
\begin{equation}
    Q_{\pi^*}(o,a) = \max_{\pi} Q_{\pi}(o,a), \quad \forall\ (o,a) \in \mathcal{O} \times \mathcal{A}.
\end{equation}
Solving the Bellman equation explicitly is only possible in very small problems, and this has led to the popularity of \gls{rl} algorithms such as \emph{Q-Learning} \cite{watkins1992q}, which iteratively updates $Q(o(k),a(k))$ with
\begin{equation}
\begin{aligned}
\label{eq:qlearn}
     Q^{(k+1)}(o(k),a(k))=&\alpha \left( r(k) + \lambda \max_a Q^{(k)}(o(k+1), a)\right) \\
&+ (1-\alpha)Q^{(k)}(o(k),a(k)),
\end{aligned}
\end{equation}
where $\alpha \in [0,1]$ is the so-called learning rate. Under most practical conditions, Q-Learning provably converges to the optimal policy if every observation-action pair is visited infinitely often. While convergence is not guaranteed, \gls{drl} algorithms, which use deep neural networks to estimate Q-values and generalize from limited experience, can deal with much more complex models and converge significantly faster. The \emph{Double Q-Learning} algorithm is an extension of \eqref{eq:qlearn} with faster convergence \cite{van2016deep}, which can also be adapted to the \gls{drl} approach. 

\subsection{Cyber-Physical POMDPs}

We have then two sets of agents, named \emph{sensors} ($\mathcal{G}$) and \emph{robots} ($\mathcal{L}$): the sensors can observe the local state of the environment and transmit their observations, while the robots have more limited observations, but can act on the environment and change its physical state.
Some agents might be able to do both and, in such cases, $\mathcal{G}$ and $\mathcal{L}$ are partially overlapping.
The environment has a set $\mathcal{S}$ of possible states, and each agent $i \in \mathcal{G} \cup \mathcal{L}$ (either sensor or robot) receives observations from a set $\mathcal{O}_i$, depending on the current state.
We denote the set of joint observations as $\mathcal{O}=\prod_{i\in\mathcal{G}\cup\mathcal{L}}\mathcal{O}_i$, such that a joint observation $o=(o_1,\ldots,o_{|\mathcal{G}|+|\mathcal{L}|})$, with $o_i\in\mathcal{O}_i$, and the probability of having a certain joint observation in a certain state as $P_o:\mathcal{S}\rightarrow\mathcal{O}$.

In a given state $s$, each agent can either communicate, or move, or take both the actions. 
Hence, the state space consists in the union of two subsets:  $\mathcal{S}_c$ and $\mathcal{S}_m$, which include \textit{communication} and \textit{movement} states, respectively.
If the environment is in a communication state, i.e., $s \in \mathcal{S}_c$, each sensor $g \in \mathcal{G}$ can transmit its information to other nodes, taking actions in the space $\mathcal{A}_{\mathcal{G}}(s)$.
For instance, $\mathcal{A}_{\mathcal{G}}(s)$ can be binary, with one action for transmission and one for silence, or have multiple values for different possible messages. 
On the other hand, if the environment is in a movement state, i.e., $s \in \mathcal{S}_m$, each robot $\ell \in \mathcal{L}$ can make an action in the physical environment, with action space $\mathcal{A}_{\mathcal{L}}(s)$.
Note that $\mathcal{S}_c\cup\mathcal{S}_m=\mathcal{S}$, which implies that at least one type of agent can make an action in each state. If $\mathcal{S}_c\cap\mathcal{S}_m\neq\emptyset$, there might be states that allow both agent types to act.

If the system is in state $s$, the joint agent actions are described by a vector $\mathbf{a} \in\ \mathcal{A}(s) = \prod_{g\in\mathcal{G}}\mathcal{A}_{\mathcal{G}}(s) \times \prod_{\ell\in\mathcal{L}}\mathcal{A}_{\mathcal{L}}(s)$, which is the concatenation of the single actions $a_i$, $\forall\ i \in \mathcal{G} \cup \mathcal{L}$. 
In particular, each agent can take actions or not depending on the current state space:
\begin{equation}
\begin{aligned}
   \mathcal{A}_{\mathcal{G}}(s)=\{a_\varnothing\},&\text{ if } s\notin\mathcal{S}_c;\\
   \mathcal{A}_{\mathcal{L}}(s)=\{a_\varnothing\},&\text{ if } s\notin\mathcal{S}_m;
\end{aligned}
\end{equation}
where $a_{\varnothing}$ is a null action with no effects.
As in a classical \gls{mdp}, the system evolution only depends on the current state and the robots' actions, and we denote the state transition probability function as $P_{s,s'}: \mathcal{S} \times\mathcal{A}_{\mathcal{G}}\rightarrow\mathcal{S}$. 

In our model, there are two different reward functions: $R_{\mathcal{G}}$ (for the sensors) and $R_{\mathcal{L}}$ (for the robots), which both have $\mathcal{S} \times \mathcal{A}$ as a domain and return values in $\mathbb{R}$.
The robots' reward is not directly dependent on the sensors: given two different action vectors $\mathbf{a}, \mathbf{b} \in \mathcal{A}$, $\mathbf{a} \neq \mathbf{b}$, the robots' reward does not change if the robots' actions are identical in the two cases.
Instead, the reward of the sensors depends on both their actions and those of the robots, as their goal is to aid the robots in their task in the physical environment.
As we are considering a cooperative scenario, in which the goal of the sensors is to help the robots in their task, the definition of a pragmatic communication entails two reward components. The first is the success of the communication, such as avoiding collisions or packet losses. The second is the \emph{relevance} of the transmitted information, i.e., how much it helps the robots improve their performance.

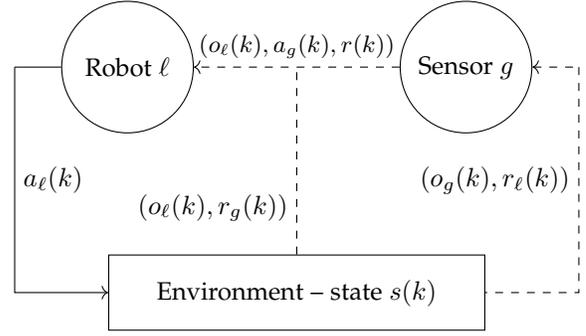
\begin{figure}[t]
    \centering
    \begin{tikzpicture}[auto]

\node[name=env,draw, minimum height=1cm, minimum width=5cm] at (0,0) {Environment -- state $s(k)$};

\node[name=a1,circle,draw,minimum width=1.75cm] at (-2.25,3) {Robot $\ell$};
\node[name=a2,circle,draw,minimum width=1.75cm] at (2.25,3) {Sensor $g$};

\draw[->,dashed] (a2.west) -- node [above, midway] {\small $(o_{\ell}(k),a_g(k),r(k))$} (a1.east);
\draw[-,dashed] (env.north) -- node [left,near start] {$(o_{\ell}(k),r_g(k))$} (0,3);

\draw[-] (a1.west) --  (-3.75,3);

 \draw[->] (-3.75,3) |- node [right,near start] {$a_{\ell}(k)$} (env.west);
 
\draw[<-,dashed] (a2.east) --  (3.75,3);

 \draw[-,dashed] (3.75,3) |- node [left,near start] {$(o_{g}(k),r_{\ell}(k))$} (env.east);

%
%
%
%
%
%
%

\end{tikzpicture}
    \caption{Simple \gls{cppomdp} model with two agents.}
    \label{fig:pomdp}
\end{figure}

A simple schematic of this model is given in Fig.~\ref{fig:pomdp}, with one sensor and one robot: while the actions of the robot affect the environment, the sensor only affects the future observations of the robot, contributing to the task indirectly by giving the robot the most useful information. The full \gls{cppomdp} model is then defined by the tuple $(\mathcal{G},\mathcal{L},\mathcal{S}_c,\mathcal{S}_m,\mathcal{A}_{\mathcal{G}},\mathcal{A}_{\mathcal{L}},P_s,P_o,R_{\mathcal{G}}, R_{\mathcal{L}})$. 
In general, the nature of the model reflects the indirect effect of the sensors, whose actions do not affect the physical state of the environment, but only the information available to other nodes.
They can then change the transition probabilities from a certain state to another that has the same physical environment, but yields richer observations to all or a subset of robots. By appropriately defining the state space and transition probabilities, it is possible to represent several different communication scenarios:
\begin{itemize}
 \item \emph{Scheduled transmission}: there is no interference between transmissions, but the set $\mathcal{A}_{\mathcal{G}}$ might be too small to encode all the information each sensor has about the system state. The sensors need to encode data by deciding which parts of their observation are more important to transmit.
 \item \emph{Slotted ALOHA}: the sensors need to coordinate to avoid interference. Each sensor can decide whether to transmit in a given state, or which resource block to use, and collisions significantly reduce the reception probability. This means that each sensor needs to estimate the usefulness of its information and make a decision on whether to transmit and risk a collision or just remain silent.
 \item \emph{Spatial reuse}: the sensors have a limited range, so concurrent transmissions can be received if the interference is low enough. Naturally, this means that direct communication between a sensor and another agent outside of its range is impossible, and opens the possibility for multi-hop transmission.
\end{itemize}

Similarly, we can define specific scenarios depending on the number of robots and the effect of their actions:
\begin{itemize}
 \item \emph{Distributed perception}: there is a single robot (i.e., $|\mathcal{L}|=1$) and all other nodes are sensors, whose objective is to augment the robot's awareness and improve its performance in the task.
 \item \emph{Coordination}: all nodes are both robots and sensors, and they must share their observations to achieve some form of consensus in order to perform an action.
 \item \emph{Consensus}: this is a special case of coordination, in which actions can only be performed if a consensus threshold is met (either full unanimity, or some form of qualified majority). A well-known example from the literature is the block pushing task for a series of puck robots, who are too weak to push the block individually but can act in concert to accomplish the task~\cite{alkilabi2017cooperative}.
\end{itemize}

\section{The Distributed Perception Case}\label{sec:dp}

Here we define an example of a distributed perception scenario, with a single robot aided by several sensors. The example will be used as a general model for the two use cases we present in the later sections of the paper.

\subsection{Use Case Scenario}\label{ssec:scenario}

We consider a gridworld scenario, in which the map is represented by a square grid of size $N \times N$. Each cell of the map is identified by a coordinate vector $\mathbf{x} \in \mathcal{M} = \mathcal{N} \times \mathcal{N}$, where $\mathcal{N} = \{1,..., N\}$. The single robot\footnote{We add the subscript $\ell$ in the notation when referring to the robot, to avoid confusion with other entities.}, whose position in slot $k$ is denoted by $\mathbf{x}_{\ell}(k)$, must reach a set $\mathcal{T}$ of  mobile targets: the position of the $i$-th target in slot $k$ is denoted by $\mathbf{t}_i(k)$. At the same time, the agent must avoid moving obstacles in the scenario. The final goal is to reach a known final position $\mathbf{x}_f$ after visiting all targets. To achieve this goal, the perception of the robot is augmented by a set of sensors $\mathcal{G}$, which can perceive their surrounding environment and communicate their observations to the robot. 

Let use denote the robot's position at slot $k$ by $\mathbf{x}_{\ell}(k)$. Following a common \gls{rl} approach, we divide the time into episodes, each of which lasts $K\in\mathbb{N}$ slots. We also define the environment matrix $\mathbf{M}_k\in\{\phi,\omega,\tau\}^{N\times N}$ over the set $\mathcal{M}$ of possible cell coordinates $\mathbf{x}$, whose elements are equal to $\phi$ if the corresponding cell is free, $\omega$ if it contains an obstacle, and $\tau$ if it contains a target. We highlight that the environment matrix is dependent on time, as obstacles and targets may move during the course of an episode.

We denote the robot's belief on the state of the environment as $\hat{\mathbf{M}}_{\ell,k}\in\{\nu,\phi,\omega,\tau\}^{N\times N}$: value $\nu$ indicates that the robot has no knowledge of the content of a cell. The robot starts with an empty belief matrix $\hat{M}_{\ell,0}(\mathbf{x})=\nu,\ \forall \mathbf{x}\in\mathcal{M}$. At the beginning of each slot $k$, it can directly sense the status of cells within a limited field of view $\mathcal{V}_{\ell}(k)$, defined as the set of cells whose Euclidean distance from the robot's current position is lower than a certain value $v_{\ell}$, and update those values in $\hat{\mathbf{M}}_{\ell,k}$. The belief matrix is then updated:
\begin{equation}
      \hat{M}_{\ell,k}(\mathbf{x}) = \begin{cases}
    M_k(\mathbf{x}), &\text{if } ||\mathbf{x}-\mathbf{x}_{\ell}(k)||_2\leq v_{\ell}; \\
    \hat{M}_{\ell,k-1}(\mathbf{x}), &\text{otherwise},
    \end{cases}
\end{equation}
where $||\cdot||_2$ denotes the Euclidean norm. We also include a matrix $\bm{\Delta}_{\ell, k}\in\mathbb{N}^{N\times N}$ in the robot's observation, which represents the \gls{aoi} of each cell, i.e., how long ago each cell was last updated:
\begin{equation}
      \Delta_{\ell,k}(\mathbf{x}) = \begin{cases}
    0, &\text{if } ||\mathbf{x}-\mathbf{x}_{\ell}(k)||_2\leq v_{\ell}; \\
    \Delta_{\ell,k-1}(\mathbf{x})+1, &\text{otherwise},
    \end{cases}
\end{equation}
with $\Delta_{\ell,0}=\infty,\ \forall \mathbf{x}\in\mathcal{M}$. The robot's full observation at step $k$ is then $o_{\ell}(k)=(\mathbf{x}_{\ell}(k),\hat{\mathbf{M}}_{\ell,k},\bm{\Delta}_{\ell, k})$. Correspondingly, the robot's observation space is $\mathcal{O}_{\ell}=\mathcal{M}\times\{\nu,\phi\omega,\tau\}^{N\times N}\times\mathbb{N}^{N\times N}$.

The robot's actions are simple, as it can move either horizontally or vertically by one cell. Hence, its action space is given by $4$ possible movements: $\mathcal{A}_{\mathcal{L}} = \{(1,0)$, $(-1,0)$, $(0,1)$, $(0,-1) \}$, corresponding to a step to the left, right, up, and down, respectively.
However, the robot can never cross the map boundaries, and obstacles are impenetrable, so if the robot tries to run into an obstacle or exit the map, its action will have no effect.
If the robot performs action $\mathbf{a}_{\ell, k} \in \mathcal{A}_{\mathcal{L}}$ at slot $k$, its position in the next slot is given by:
\begin{equation}
\label{eq:robot_action}
    \mathbf{x}_{\ell, k+1} =
    \begin{cases}
    \mathbf{x}_{\ell, k}, & \text{if } \mathbf{x}_{\ell, k} + \mathbf{a}_{\ell, k} \notin\mathcal{M};\\
    \mathbf{x}_{\ell, k}, & \text{if } M_k(\mathbf{x}_{\ell, k} + \mathbf{a}_{\ell, k})=\omega;\\
    \mathbf{x}_{\ell, k} + \mathbf{a}_{\ell, k}, & \text{otherwise.}
    \end{cases}
\end{equation}
The system is always in a movement state (i.e., $\mathcal{S}_m = \mathcal{S}$), which implies that the robot performs a new action $\mathbf{a}_{\ell, k}$ at each slot $k$.

Regarding sensors, the communication channel is severely constrained, such that the system is in a communication state $s \in \mathcal{S}_c$ only when $\text{mod}(k,K_p) = 0$, where $\text{mod}(\cdot)$ is the integer modulo function; that is, sensor actions are separated by $K_p$ slots. As all slots are movement slots, both the sensors and the robot will act in the communication slots: the sensors will transmit their information, and the robot will move in the grid.

Like the robot, sensors also have a limited field of view, with a maximum distance $v_g$ for each sensor $g$. However, the position of each sensor $g$, $\mathbf{x}_g$, is fixed. We can then define sensor $g$'s field of view $\mathcal{V}_g$:
\begin{equation}
 \mathcal{V}_g=\left\{\mathbf{x}\in\mathcal{M}:||\mathbf{x}-\mathbf{x}_g||_2\leq v_g\right\}.
\end{equation}
Sensors can transmit its information over a subset $\mathcal{C}_g\subseteq\mathcal{V}_g$ of the map, corresponding to a limited area of the map: each sensor's action is then its choice of $\mathcal{C}_g$. After a successful communication, the robot's new belief matrix $\hat{M}'_{\ell,k}(\mathbf{x})$ is then given by:
\begin{equation}
      \hat{M}'_{\ell,k}(\mathbf{x}) = \begin{cases}
    M_k(\mathbf{x}), &\text{if } \mathbf{x}\in\mathcal{C}; \\
    \hat{M}_{\ell,k}(\mathbf{x}), &\text{otherwise},
    \end{cases}
\end{equation}
The value of $\bm{\Delta}'_{\ell,k}$ is also reset to 0 for all cells in the transmitted area $\mathcal{C}_g$.
We define two cases, corresponding to different communication scenarios:
\begin{itemize}
    \item In the centralized control scenario, a single sensor $g$ (with $|\mathcal{G}|=1$) can perceive the full map, i.e., $v_g=\infty$ and $\mathcal{V}_g=\mathcal{M}$. In this case, the channel capacity is the constraining factor, so that the sensor needs to select an area $\mathcal{C}_{i} \in \mathcal{C}$, where $\mathcal{C}$ is a partition of the map.
    Therefore, $\mathcal{A}_{\mathcal{G}} = \mathcal{C}$. This configuration is easy to manage, as communication decisions are centralized and communication is always successful.
    \item In the distributed control scenario, there are multiple sensors, i.e., $|\mathcal{G}|>1$, that can transmit over a shared collision channel. If the system is in a communication state, each sensor can independently decide whether to transmit information about its field of view (action 1) or remain silent (action 0); therefore, each sensor's action space is $\mathcal{A}_{\mathcal{G}} = \{0,1\}$. If the sensor chooses action 1 and transmits, $\mathcal{C}_g=\mathcal{V}_g$. However, since the transmission is over a collision channel, the transmission is only successful if $\sum_{g\in\mathcal{G}}a_{g}=1$. Since the sensors act simultaneously without any explicit coordination, it is challenging to find a communication policy that allows the robot to receive enough information while minimizing the risk of packet collisions.
    In particular, each sensor is not aware of the information observed by the others, and needs to consider whether its observations are important enough to transmit and risk a collision or whether some other sensor may have more important information.
\end{itemize}

The ultimate goal of the system is for the robot to accomplish its task in as few steps as possible; as such, a large reward $\rho$ for the robot is given when it reaches the target position (no reward is given if the episode ends). In order to accelerate the training and make the task easier, a smaller intermediate reward $\sigma$ might be given if a condition is met, which we denote as the generic indicator function $\chi(s(k),a_{\ell})$. The reward for the robot, $r_{\ell}(k)$, is then:
\begin{equation}
r_{\ell}(k) =
    \begin{cases}
    \rho, \quad &\text{if } \mathbf{x}_{\ell}(k)= \mathbf{x}_{f}\wedge\forall t\in\mathcal{T},\,\exists j:\mathbf{x}_{\ell}(j)=\mathbf{x}_t(j) ; \\
    \sigma, \quad &\text{if } \chi(s(k), \mathbf{a}_{\ell}(k))=1; \\
    0, \quad & \text{otherwise}.
    \end{cases}\label{eq:reward}
\end{equation}
On the other hand, the reward for each sensor $g$ is given by the reward for the robot over the next time horizon, with an additional penalty $\eta$ for collisions in the distributed case:
\begin{equation}
        r_{g}(k) = \begin{cases}
    -\eta, & \text{if } a_{g}(k)\neq0\wedge\sum_{j\in\mathcal{G}}a_{j}(k)\geq 2; \\
    \sum_{j=k} ^{k + K_p} r_{\ell}(j), &\text{otherwise.}
    \end{cases}
\end{equation}
Naturally, there is no penalty in the centralized case, as the sensor does not suffer from any interference and communication is ideal.

\subsection{Learning Architecture}\label{ssec:learning}

To determine the best policy for managing our system, we exploit the \gls{dqn} architecture~\cite{mnih2015human}, which approximates function $Q(\cdot)$ using a \gls{nn}.
In this case, we need to learn multiple functions, one for each agent in $\mathcal{L} \cup \mathcal{G}$.
In particular, in the centralized control scenario, we define the functions $Q_{\ell}(\cdot)$ and $Q_{g}(\cdot)$, which determine the movements of the robot and the sensor's communication strategy, respectively.
On the other hand, in the distributed control scenario, we have a total of $1 + | \mathcal{G}|$ quality functions, i.e., one managing the robot, and an additional function for each of the sensors.   

In the case of the robot, the function $Q(\cdot)$ is approximated by a \gls{cnn} with $4$ output neurons, i.e., one for each possible movement.
The \gls{cnn} input is a convolutional layer that handles the robot's observation as an $N \times N$ image with 4 channels.  
The input channels are given by the matrices $\hat{\mathbf{M}}_{\ell,k}$, and $\bm{\Delta}_{\ell,k}$, as well as the one-hot encoding matrices $\mathbf{X}_{\ell,k}$ and $\mathbf{X}_{f}$, which represent the encodings of $\mathbf{x}_{\ell}(k)$ and $\mathbf{x}_f$, respectively. Therefore, at each slot $k$, the neural network takes these 4 matrices as its input information and returns the expected value of each action $a \in \mathcal{A}_{\mathcal{L}}$ as outputs.

The function $Q(\cdot)$ of each sensor $g \in \mathcal{G}$ is instead approximated by a \gls{cnn} using a convolutional layer with 5 channels as input, and a linear layer with $|\mathcal{A}_{\mathcal{G}}|$ neurons as output.
In this case, each observation is treated as an $N \times N$ image with 5 channels: the first 4 channels encode the same information as the robot, while the fifth encodes matrix $\hat{M}_{g,k}$, which is defined as follows:
\begin{equation}
    \hat{M}_{g,k}(\mathbf{x}) = \begin{cases}
    M_k(\mathbf{x}), &\text{if }\mathbf{x}\in\mathcal{V}_g; \\
    \nu, &\text{otherwise.}
    \end{cases}
\end{equation}
The outputs are 2 in the distributed scenario, and $|\mathcal{C}|$ in the centralized scenario.

The network architectures for the robot and sensors all have four convolutional layers followed by two fully connected linear layers. We consider the \gls{relu} activation function for the hidden layers because of its efficiency in representing non-linear dynamics, while we implement a linear function in the output layer.  

\subsection{Training Framework} \label{sec:training}

In order to achieve a \gls{jcc} optimization, in which the robot and sensors' policies are optimzied for each other. Both the robot and the sensors are controlled by \gls{drl} agents, and the system training follows an iterative approach. This is a common practice in \gls{marl}, as multiple agents exploring the environment and improving their policies at the same time can prevent convergence: by only training one kind of agent at a time, we ensure that the environment that agent sees is Markovian.

The training consists in $N_{\text{round}}$ rounds, each of which includes two phases: the first lasts $N_{\text{train}}^{\ell}$ episodes and is dedicated to the robot training, while the second lasts $N_{\text{train}}^{g}$ episodes, and is dedicated to the sensor training. In the first phase, the policy of the sensors is fixed (in the first round, the sensors are initialized with a pre-determined strategy, while in successive rounds, the policy results from the outcome of the previous round. The opposite happens in the second phase, in which the robot's \gls{dqn} is frozen, while the sensors are trained. Once this iterative training phase is over, we train the robot architecture again for $N_{\text{train}}^{\ell}$ episodes.
Hence, the total number of robot training episodes is $N_{\text{train}}^{\ell} \times (N_{\text{round}} + 1)$, while the sensors are trained for $N_{\text{train}}^{g} \times N_{\text{round}}$ episodes. 
The described framework makes it possible to fully explore the learning environment, while ensuring sufficient robustness in the training phase.

\begin{table*}[t!]
\centering
\caption{Agent architecture.}
\label{tab:learn_param}
\begin{tabular}{@{}c|llllllll@{}}
\toprule
& \multicolumn{8}{c}{Layer} \\
\cmidrule(l){2-9}
\multirow{-2}{*}{Agent} & \multicolumn{2}{l}{Input (convolutional)} & \multicolumn{2}{l}{Hidden (convolutional)} & \multicolumn{2}{l}{Hidden (convolutional)} & \multicolumn{2}{l}{Output (linear)} \\
\midrule
& {Input channel} & 4 & {Input channel} & 64 & Input channel & 64 & &  \\
& {Output channel} & 64 & {Output channel} & 64 & Output channel & 64 & \multirow{-2}{*}{Input neurons} & \multirow{-2}{*}{64}  \\
& Kernel & (3,3) & {Kernel} & (3,3) & {Kernel} & (3,3) & & \\
\multirow{-4}{*}{AUV} & Stride & 1 & Stride & (4,4) & Stride & (4,4) & \multirow{-2}{*}{Output neurons} & \multirow{-2}{*}{4} \\
\midrule
& {Input channel} & 4 & {Input channel} & 64 & Input channel & 64 & &  \\
& {Output channel} & 64 & {Output channel} & 64 & Output channel & 64 & \multirow{-2}{*}{Input neurons} & \multirow{-2}{*}{64}  \\
& Kernel & (3,3) & {Kernel} & (3,3) & {Kernel} & (3,3) & & \\
\multirow{-4}{*}{Buoy} & Stride & 1 & Stride & (4,4) & Stride & (4,4) & \multirow{-2}{*}{Output neurons} & \multirow{-2}{*}{$|\mathcal{A}_{\mathcal{G}}|$} \\
\bottomrule
\end{tabular}
\end{table*}

\begin{table}[t!]
\centering
\caption{Simulation settings.}
\label{tab:sim_param}
\begin{tabular}{@{}lll@{}}
\toprule
Parameter & Value & Description  \\
\midrule
$|\mathcal{G}|$ & 9 & Number of buoys  \\
$N \times N$ & $12 \times 12$ & Map size \\
$C$ & 9 & Number of transmission areas \\
$K$ & 100 & Step per episode \\
$K_p$ & 5 & Inter-transmission period  \\
$\rho$ & 1.0 & Final reward \\
$\sigma$ & 0.22 & Intermediate reward \\
$\lambda$ & 0.95 & Discount factor \\
$\zeta$ & 0.00001 & Learning rate \\
$N_{\text{round}}$ & 2 & Training rounds \\
$N_{\text{train}}^{\ell}$ & 100000 & Training episodes for the \gls{auv} \\ $N_{\text{train}}^{g}$ & 500000 & Training episodes for the buoys \\
$N_{\text{test}}$ & 10000 & Testing episodes \\
\bottomrule
\end{tabular}
\end{table}

\section{Simulation Settings and Results}\label{sec:analysis}

In this section, we implement our \gls{cppomdp} model in two underwater use cases, which reflect the distributed perception model from Sec.~\ref{sec:dp}. In these tasks, an \gls{auv} (which acts as the single robot in our model) must reach a surface vessel on the other side of a map, while either avoiding underwater debris or visiting a set of nodes to recover data. A set of buoys act as sensors; in the centralized case, the buoys also have wireless connections over the air, and so can act as a single entity and share information, while in the distributed case, they must use a shared underwater acoustic channel and risk packet collisions.

First, we describe the benchmark strategies against which our learning framework is tested and present the simulation settings.
Finally, we analyze the system performance in the two test scenarios, highlighting the benefits and drawbacks of the proposed \gls{jcc} approach. The settings of the learning architectures are summarized in Tab.~\ref{tab:learn_param}.

\subsection{Benchmark Strategies} \label{sec:benchmark}

As benchmarks for our model, we consider three learning systems trained according to a \gls{nc} approach: in this paradigm, the sensors follow a pre-determined communication strategy (which is given below for each benchmark system), while the \gls{auv} adapts to the environment and learns to act with a fixed communication policy.
In other words, the \gls{auv} considers the buoys as part of the learning environment and adapts its policy to the buoys' behavior. Hence, the learning environment is more stable, but the final performance is constrained by the buoys' initial policy. We implement three different algorithms to manage the communication process of the buoys. Only the \gls{auv}'s \gls{nn} is trained for a total number of $N_{\text{train}}$ episodes.

In practice, at each communication slot, the benchmarks manage the buoy's transmissions as follows:
\begin{itemize}
    \item \gls{rc}: the buoy transmits a random portion of the map $\mathcal{C}_i \in \mathcal{C}$;
    \item \gls{cc}: the buoy transmits the map sub-area $\mathcal{C}_i$ which is closest to the \gls{auv};
    \item \gls{oc}: the buoy transmits the entire map, so that $\hat{M}_{\ell, k} = M_{k}$, $\forall$ $k$.
\end{itemize}
We highlight that the \gls{oc} strategy represents an upper bound for the performance of our system, as it models a case in which there is no communication bottleneck and the \gls{auv} is always aware of the real map state.

The above strategies are tested against two configurations of our \gls{jcc} model.
The first, named \gls{cjcc}, is based on the centralized control system defined in Sec.~\ref{ssec:scenario}, while the other, named \gls{djcc}, makes use of multiple buoys, as in the distributed control scenario given in the same section. We observe that the \gls{djcc} approach is more challenging, since buoys communicate independently and packet collisions are possible.

\begin{figure*}[t]
\centering
    \subfloat[%
  Debris avoidance. \label{fig:debris}%
]{\includegraphics[width=0.45\linewidth]{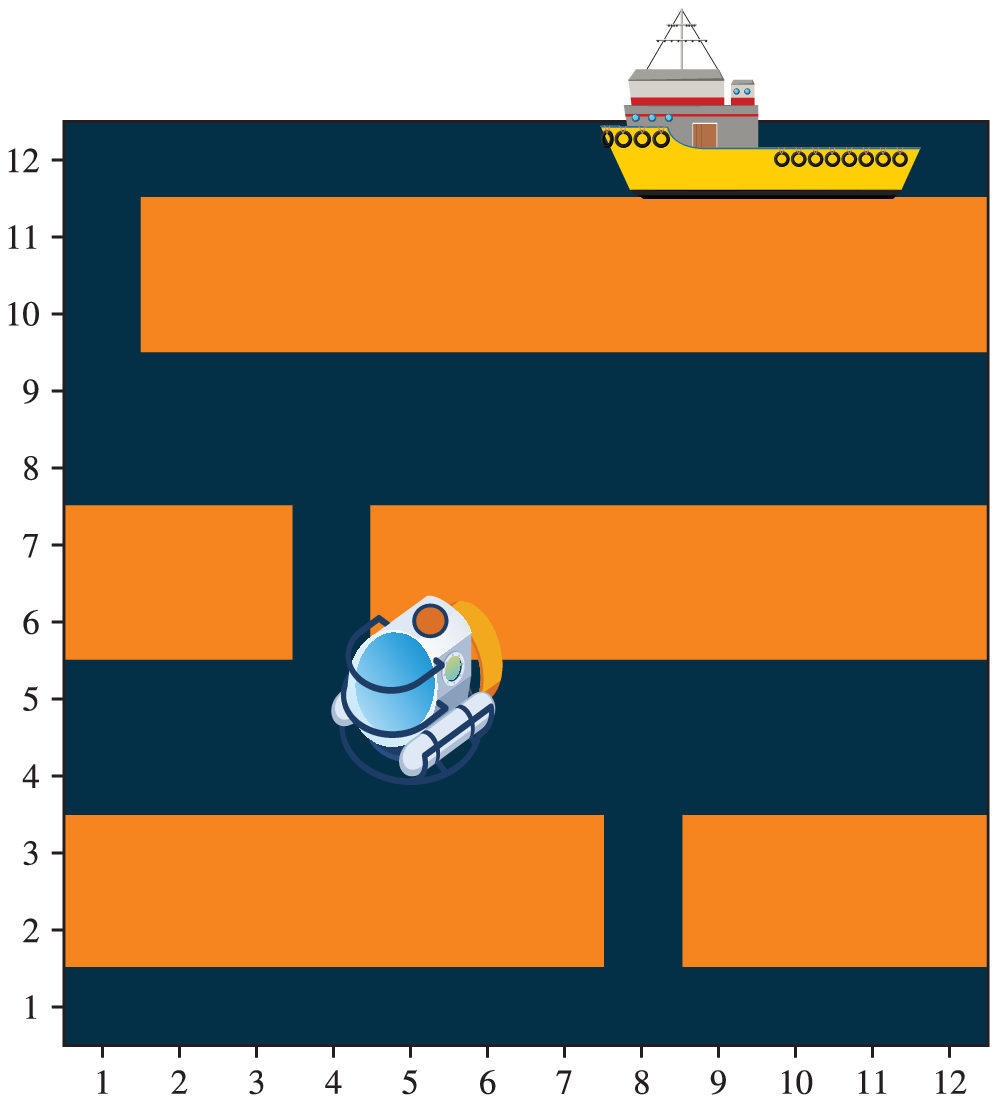}}
    \subfloat[%
  Data muling. \label{fig:muling}%
]{\includegraphics[width=0.45\linewidth]{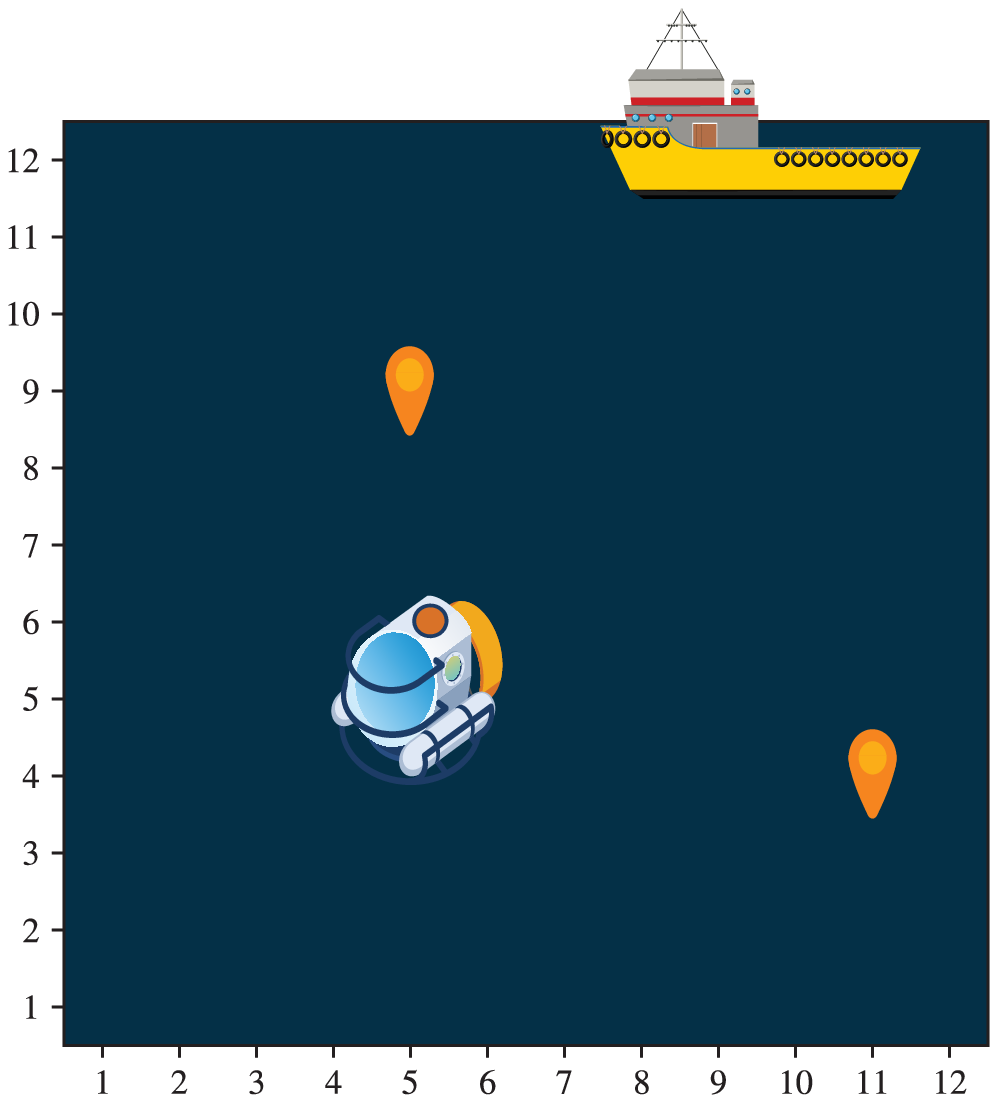}}
    \caption{Depiction of the two underwater tasks.}
    \label{fig:scenarios}
\end{figure*}

\subsection{Underwater Tasks and Scenario Settings} \label{sec:setting}
We can now define the two underwater tasks which we use for our performance evaluation. 
In both cases, we consider the reference model described in Sec.~\ref{sec:dp}, with a map of size $N=12$, divided into $T=9$ different non-overlapping portions. 
Hence, in the centralized control scenario (used by the benchmarks and the \gls{cjcc} system), there is a single buoy with $9$ possible actions. 
Instead, \gls{djcc} considers $9$ different buoys, each of which can perceive and transmit a specific portion of the map. 
We set the period between two consecutive communication states to $K_p=5$ slots and the maximum episode duration to $K=100$ slots.
If the \gls{auv} spends more than $100$ slots in the map without finishing the task, it fails to achieve its goal and the episode ends.
If the \gls{auv} reaches the surface vessel with the data, it receives a reward $\rho=10$, while the collision penalty parameter is set to $\eta=1$. 
The intermediate reward $\sigma$ instead varies over time and increases as the \gls{auv} approaches the end of its mission. 

When training the benchmark strategies, we consider a unique training phase of $N_{\text{train}} = 100000$ episodes. 
On the other hand, the training of the \gls{jcc} systems includes $N_{\text{round}} = 2$ rounds, where the \gls{auv} training phases last $N_{\text{train}}^g=100000$ episodes each, and those of the buoys $N_{\text{train}}^g=500000$ episodes each. 
We observe that $N_{\text{train}}^g$ is higher than $N_{\text{train}}^{\ell}$, since the buoys take actions less frequently than the \gls{auv} and, consequently, need more episodes to reach convergence.
During a single training round, the agents use an \textit{$\varepsilon$-greedy policy} to explore the observation-action space. The policy selects the best possible action with probability $1-\varepsilon$ and a random suboptimal action with probability $\varepsilon$, with the value of $\varepsilon$ slowly decreasing from $0.9$ to $0.1$.
Finally, we set the discount factor to $\lambda=0.95$ and we exploit the \gls{adam} algorithm to optimize the \gls{nn} architectures~\cite{kingma2017adam}, considering $\zeta=0.00001$ as the maximum learning rate.
All the simulation parameters are reported in Tab.\ref{tab:sim_param}. 

In both tasks, the final position corresponds to the position of a surface vessel on the top row of the map, which is fixed for the duration of an episode but may be different for each episode. The initial position of the \gls{auv} is also randomly generated on the bottom row of the map. Fig.~\ref{fig:scenarios} gives a visual representation of the two tasks, which are described below.

\subsubsection{Debris Avoidance}

In this scenario, there are several obstacles in the marine environment, such as rocks, sea mines, or oil installations, and the \gls{auv} must reach the vessel as fast as possible without crashing into them. 
In general, the debris locations change in time (accounting for currents and drift), so they correspond to the obstacles in $\mathbf{M}_k$. There are no targets in this environment, i.e., $\mathcal{T}=\varnothing$.

In general, we model the obstacles as 3 horizontal walls, each of which has a single opening that may move over time. The intermediate reward is given to the robot if it moves upward in the map, or if it moves sideways toward an opening when the way upward is blocked. For the full definition of the obstacles' initial positions and further movements, as well as a rigorous definition of the intermediate reward indicator function $\chi$, we refer the reader to the Appendix.

\subsubsection{Data Muling}

In the data muling scenario, the \gls{auv} has to visit a set of underwater floating nodes to recover their data through high-throughput short-range optical communications~\cite{doniec2013autonomous} before reaching the vessel to upload the data. We consider a simple case with only 2 targets, i.e., $|\mathcal{T}|=2$. In this task, we do not consider any obstacles, i.e., $\nexists\mathbf{x}\in\mathcal{M}:M_k(\mathbf{x})=\omega,\,\forall k\in\mathbb{N}$.

For the full definition of the targets' initial positions and further movements, as well as the definition of the intermediate reward indicator function $\chi$, we refer the reader to the Appendix.

\begin{figure*}[t]
\centering
    \subfloat[%
  \gls{cc} strategy. \label{fig:training_cc}%
]{\includegraphics[width=0.45\linewidth]{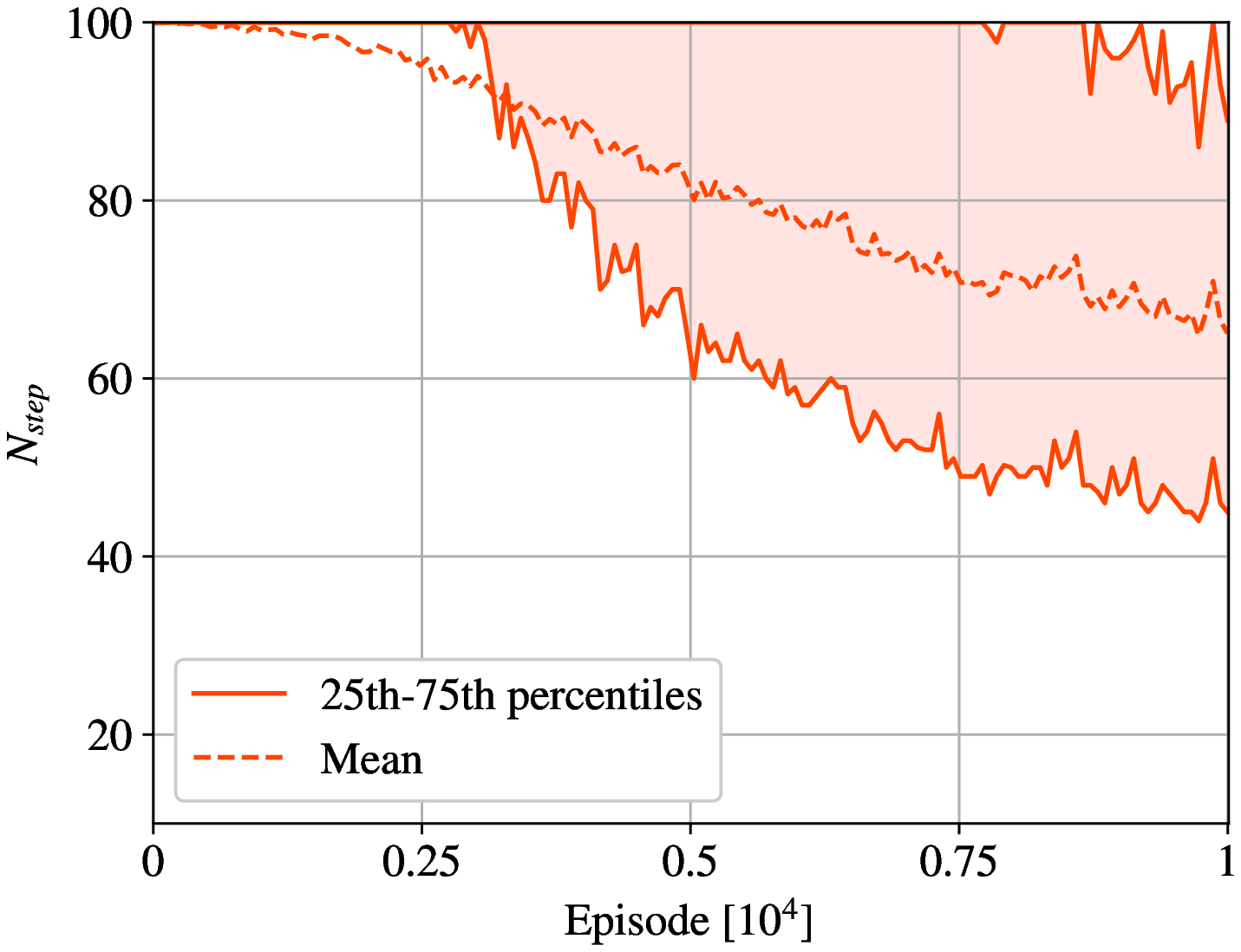}}
    \subfloat[%
  \gls{cjcc} strategy. \label{fig:training_jccc}%
]{\includegraphics[width=0.45\linewidth]{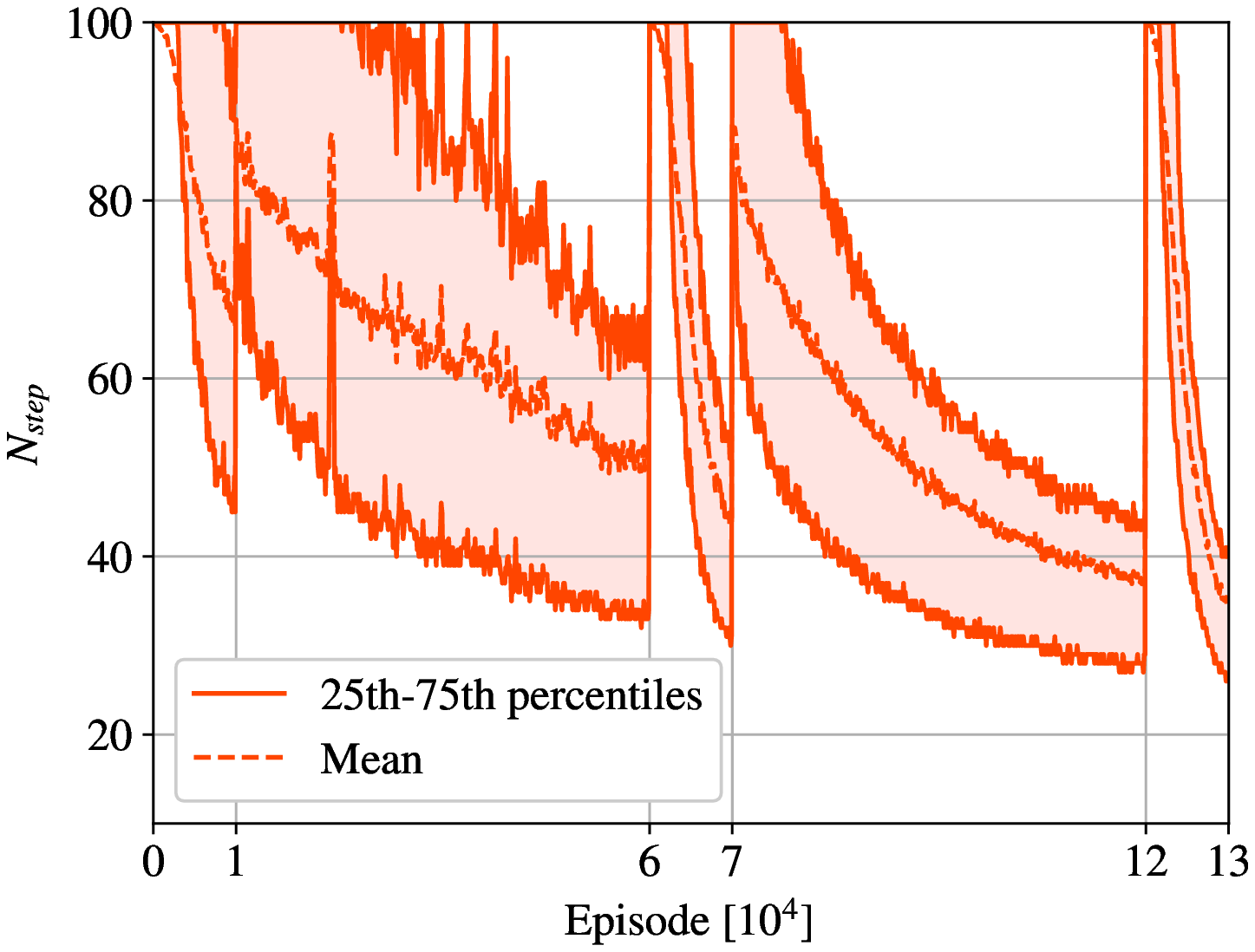}}
    \caption{Training phase (debris avoidance).}
    \label{fig:training}
\end{figure*}

\begin{figure*}[t]
\centering
    \subfloat[%
  Debris avoidance. \label{fig:perf_debris}%
]{\includegraphics[width=0.45\linewidth]{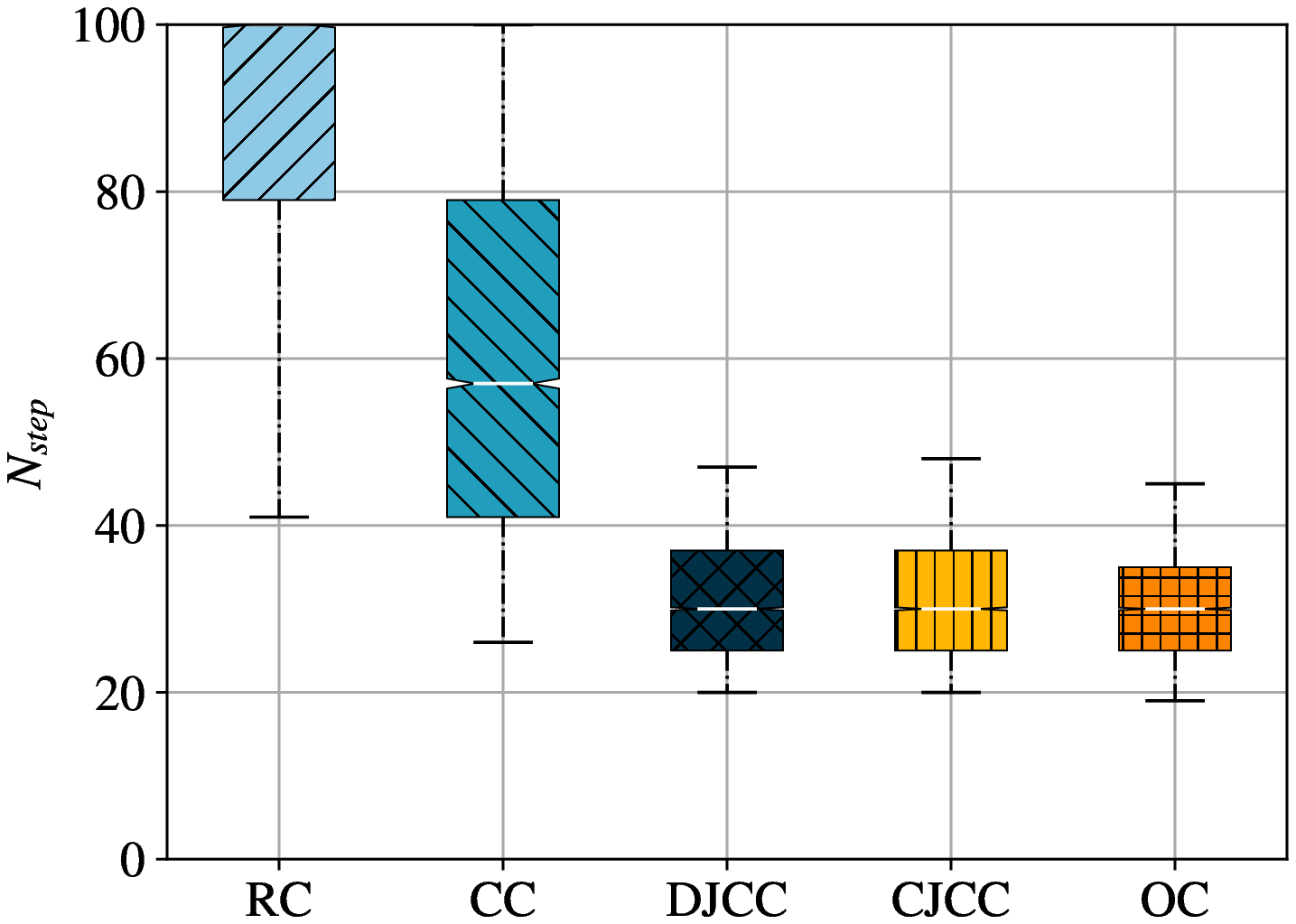}}
    \subfloat[%
  Data muling. \label{fig:perf_muling}%
]{\includegraphics[width=0.45\linewidth]{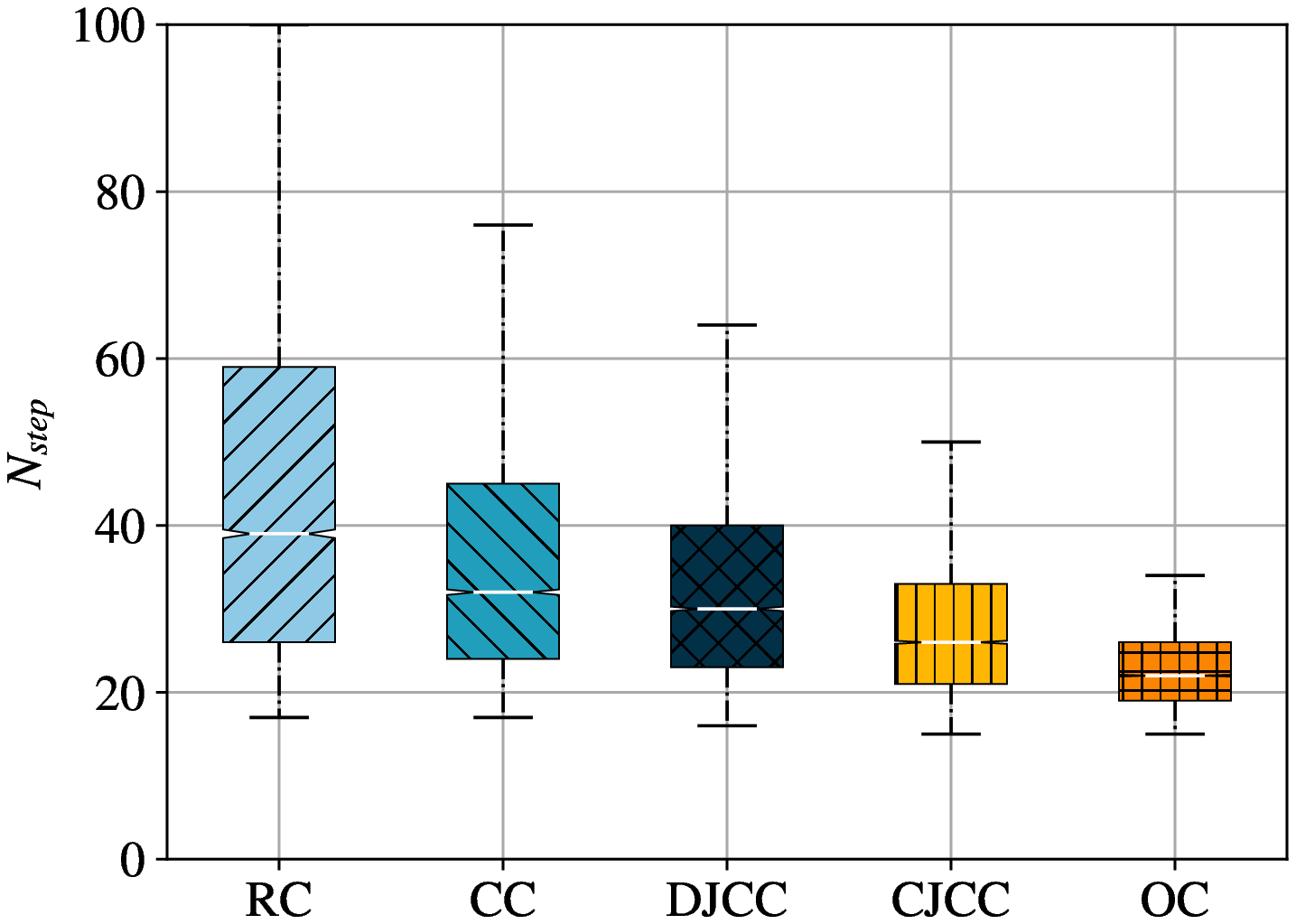}}
    \caption{Performance of the different strategies for the two tasks.}
    \label{fig:performance}
\end{figure*}

\subsection{Results} \label{sec:results}

As the main performance metric of our experiments, we consider the number $N_{\text{step}}$ of steps that the \gls{auv} takes to perform its task. 
Particularly, if the \gls{auv} is not able to achieve its goal before the episode ends, we have that $N_{\text{step}}=K$, which is the maximum episode duration.

Fig.~\ref{fig:training} shows the mean of $N_{\text{step}}$, as well as the 25th and 75th percentiles of its distribution, during the training of the \gls{cc} (above) and \gls{cjcc} (below) strategies in the ostacle avoidance scenario. 
The mean value and quartiles of $N_{\text{step}}$ decrease almost monotonically for \gls{cc}.
In \gls{cjcc}, the distribution of $N_{\text{step}}$ exhibits a regular pattern, where the average of $N_{\text{step}}$ repeatedly drops from $100$ to about half of that value.
This is because the \glspl{nn} are periodically reset after each round of the iterative training, making the \gls{auv} and the buoys forget the policy learned in the past. 
At the same time, as more training rounds are completed, the average values of $N_{\text{step}}$ become significantly lower, since the agents have progressively learned to adapt to each other. 

We carry out additional evaluation with $N_{\text{test}}=10000$ episodes for each strategy.
The distribution of $N_{\text{step}}$ during the test phase for the two considered \gls{auv} tasks is presented as a boxplot in Fig.~\ref{fig:performance}. The whiskers represent the $5$-th and $95$-th percentiles of the distribution, the edges the $25$-th and the $75$-th percentiles, and the white line in the middle of each box is the distribution median.

\begin{figure*}[t]
\centering
    \subfloat[%
  Robot location (\gls{cc}). \label{fig:location_cc_1}%
]{\includegraphics[width=0.24\linewidth]{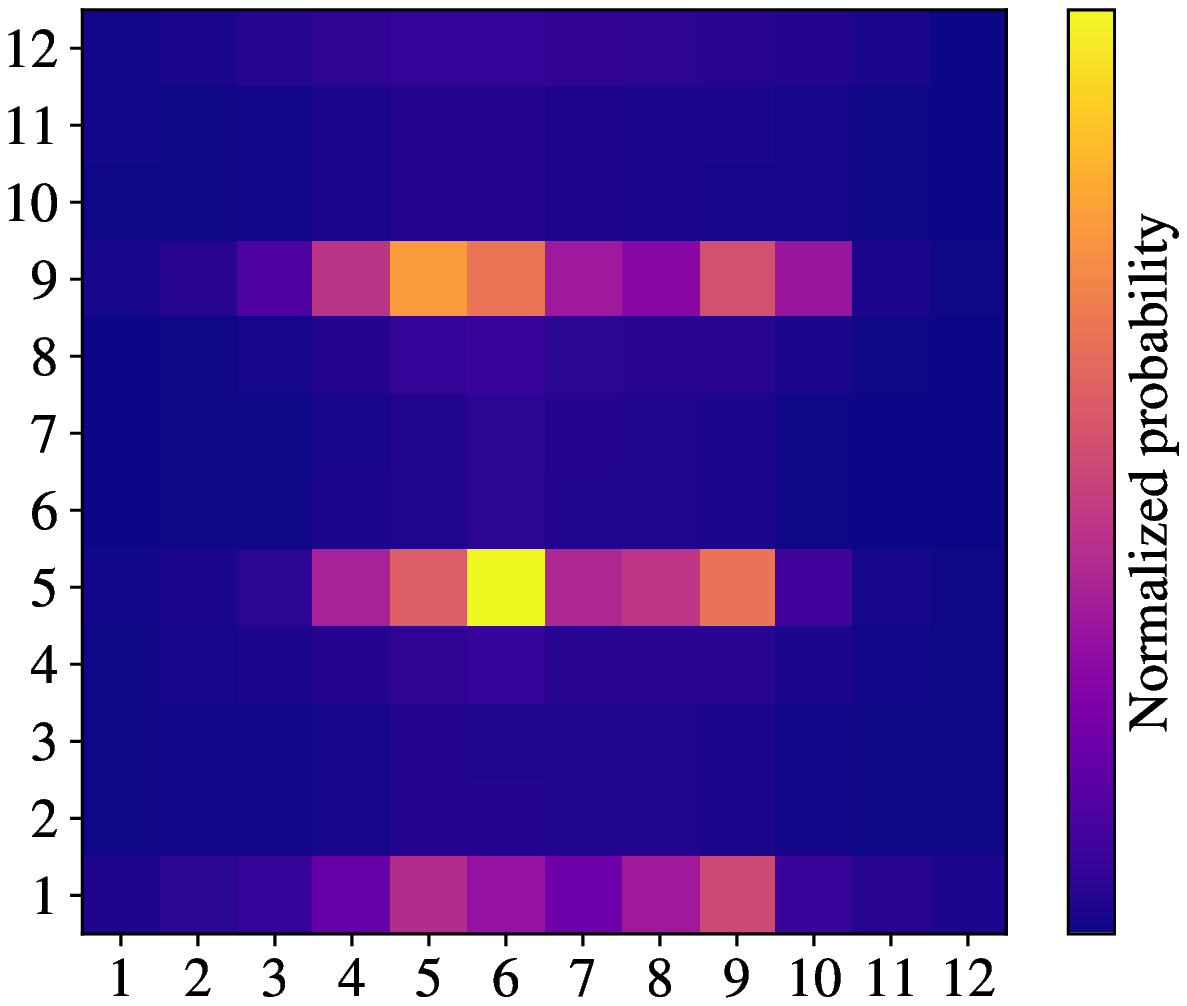}}
    \subfloat[%
  Robot location (\gls{djcc}). \label{fig:location_djcc_1}%
]{\includegraphics[width=0.24\linewidth]{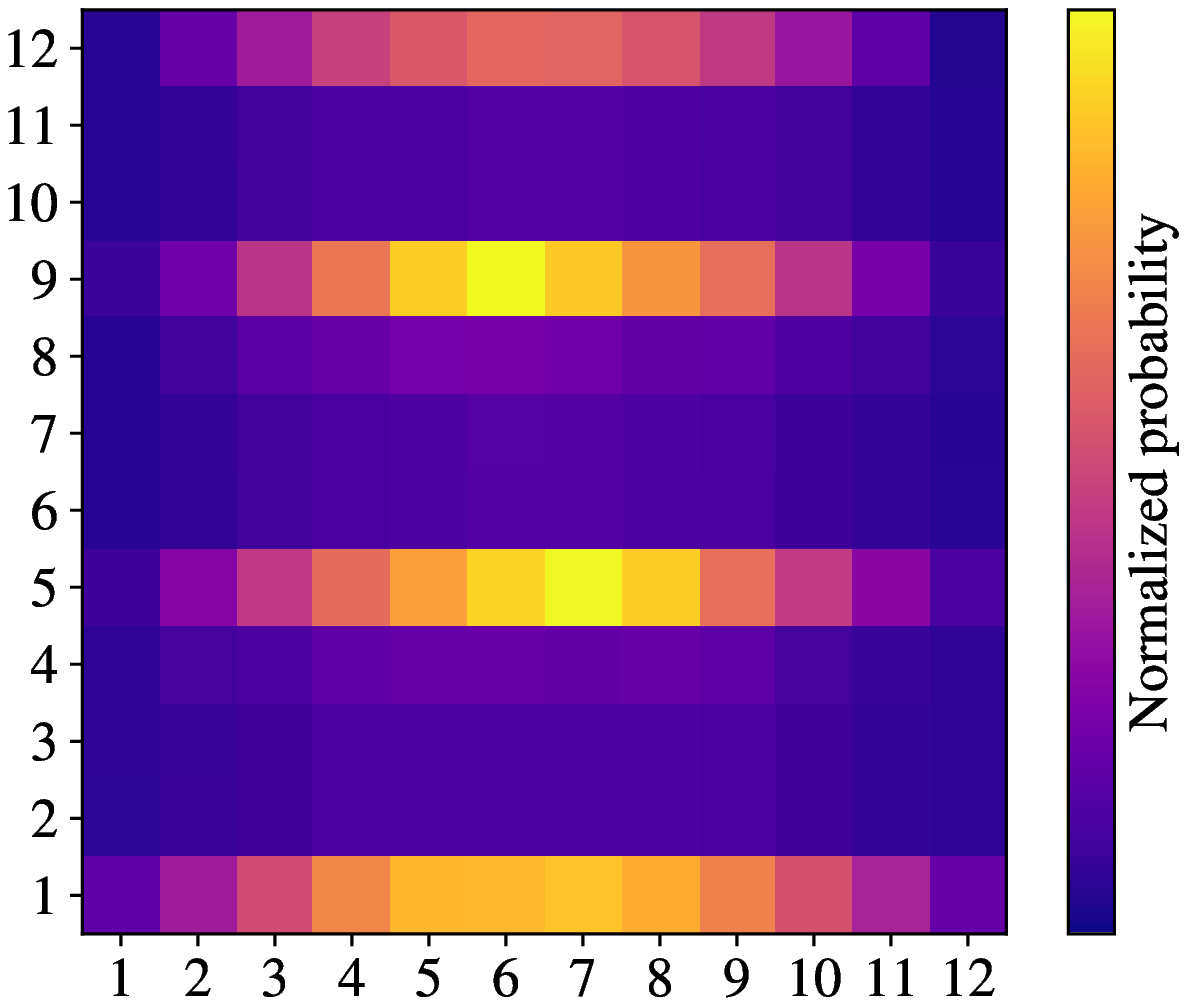}}
    \subfloat[%
  Robot location (\gls{cjcc}). \label{fig:location_cjcc_1}%
]{\includegraphics[width=0.24\linewidth]{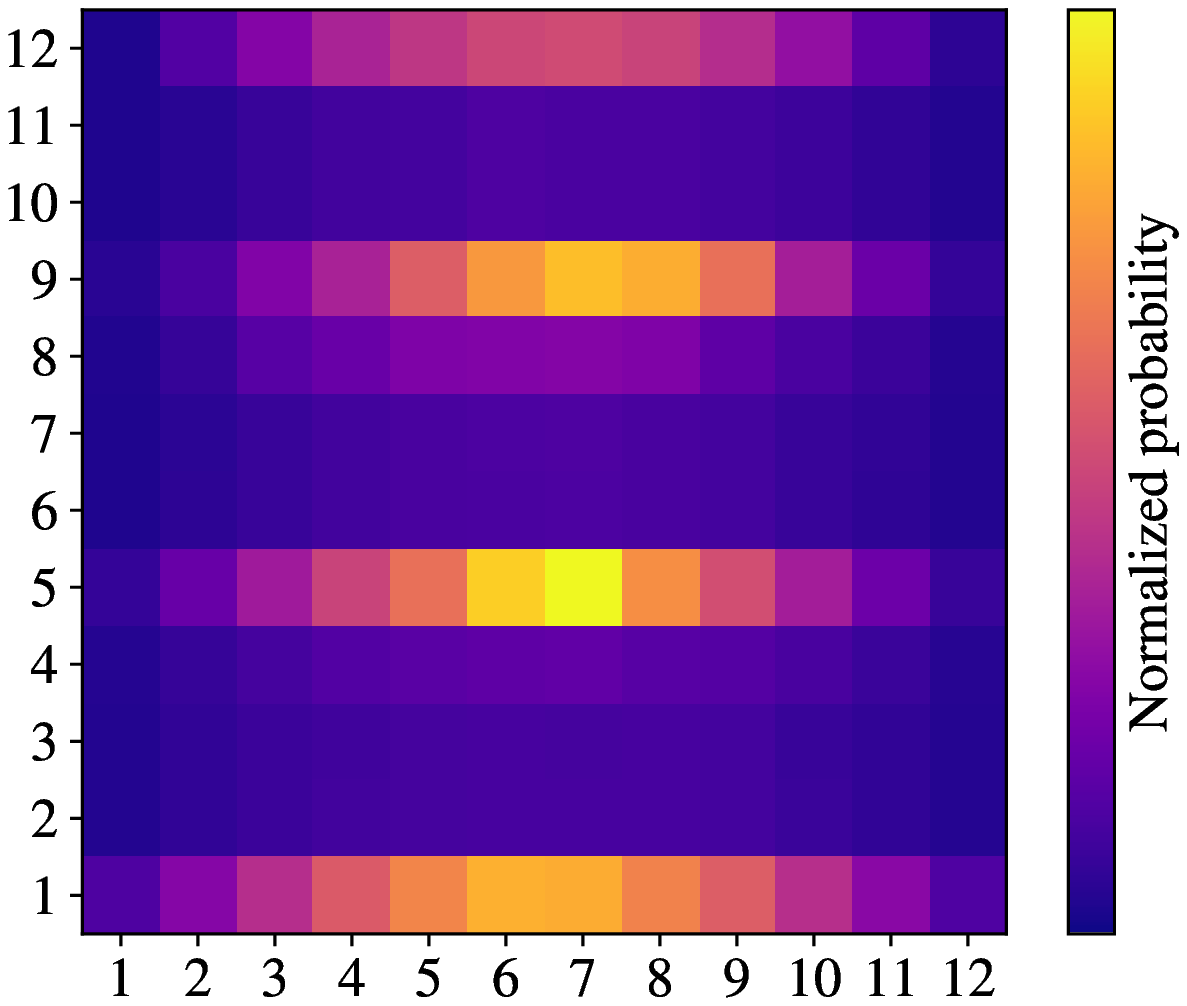}}
    \subfloat[%
  Robot location (\gls{oc}). \label{fig:location_oc_1}%
]{\includegraphics[width=0.24\linewidth]{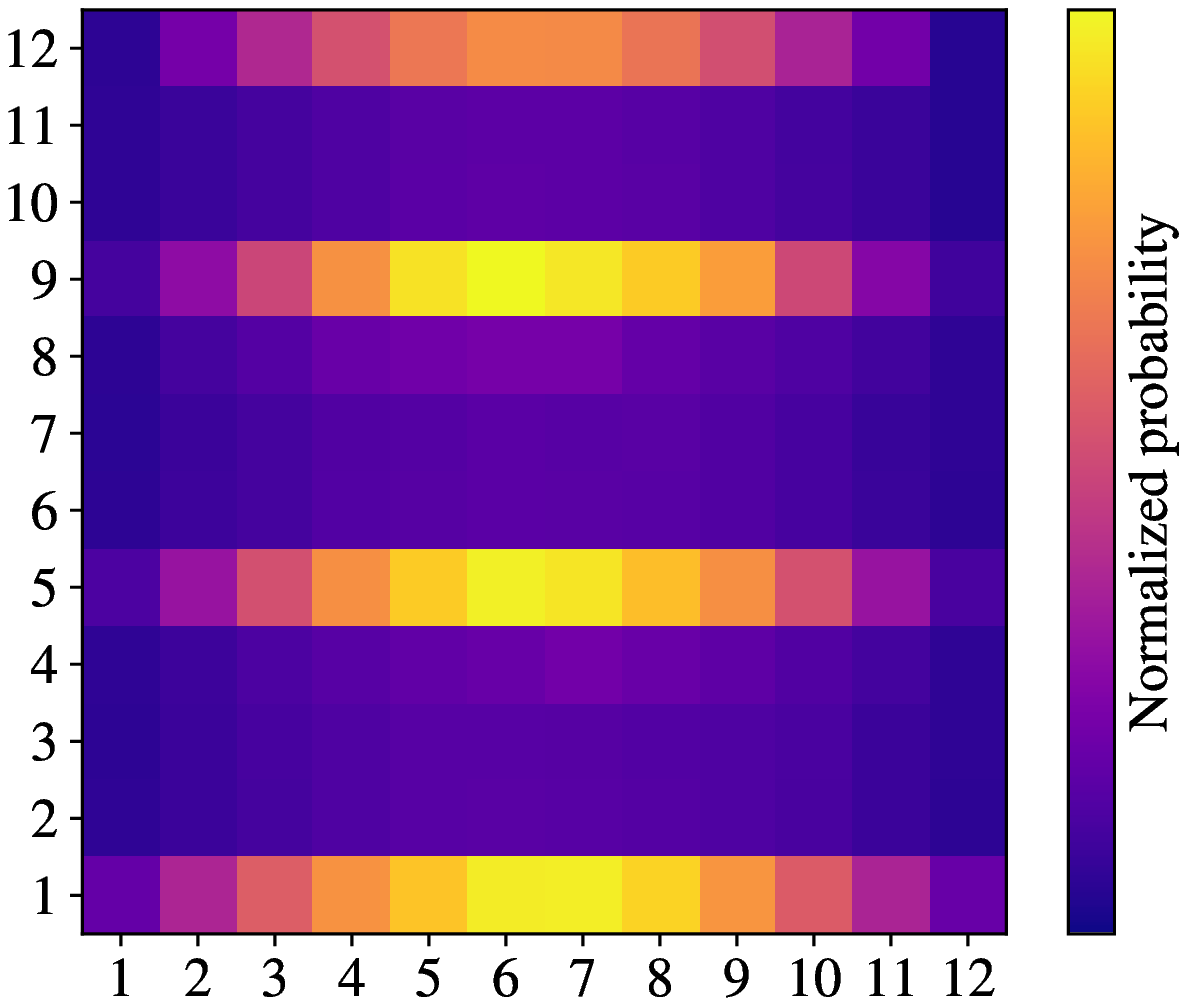}}\\
    \subfloat[%
  Sensor transmission (\gls{cc}). \label{fig:transmission_cc_1}%
]{\includegraphics[width=0.24\linewidth]{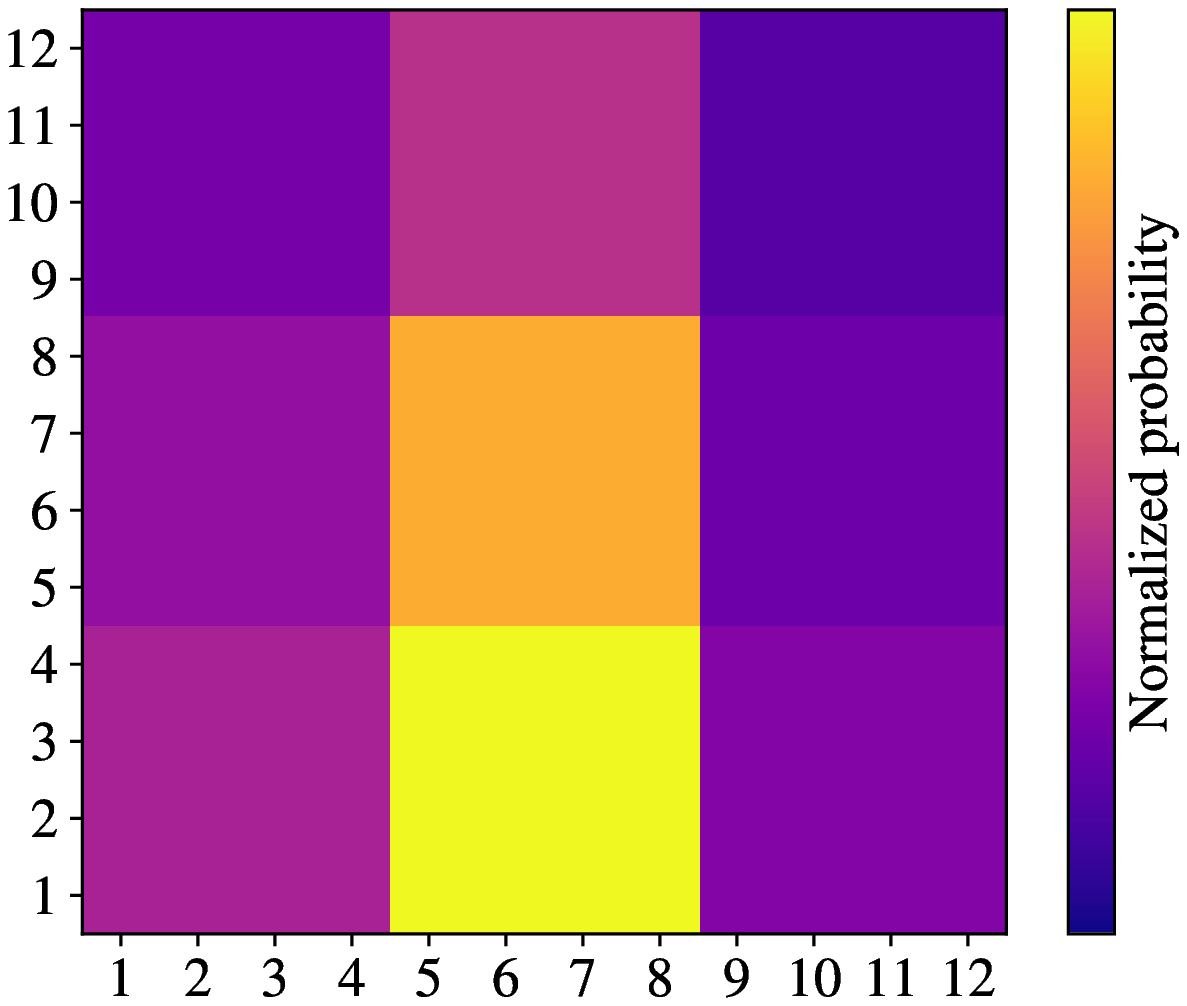}}
    \subfloat[%
  Sensor transmission (\gls{djcc}). \label{fig:transmission_djcc_1}%
]{\includegraphics[width=0.24\linewidth]{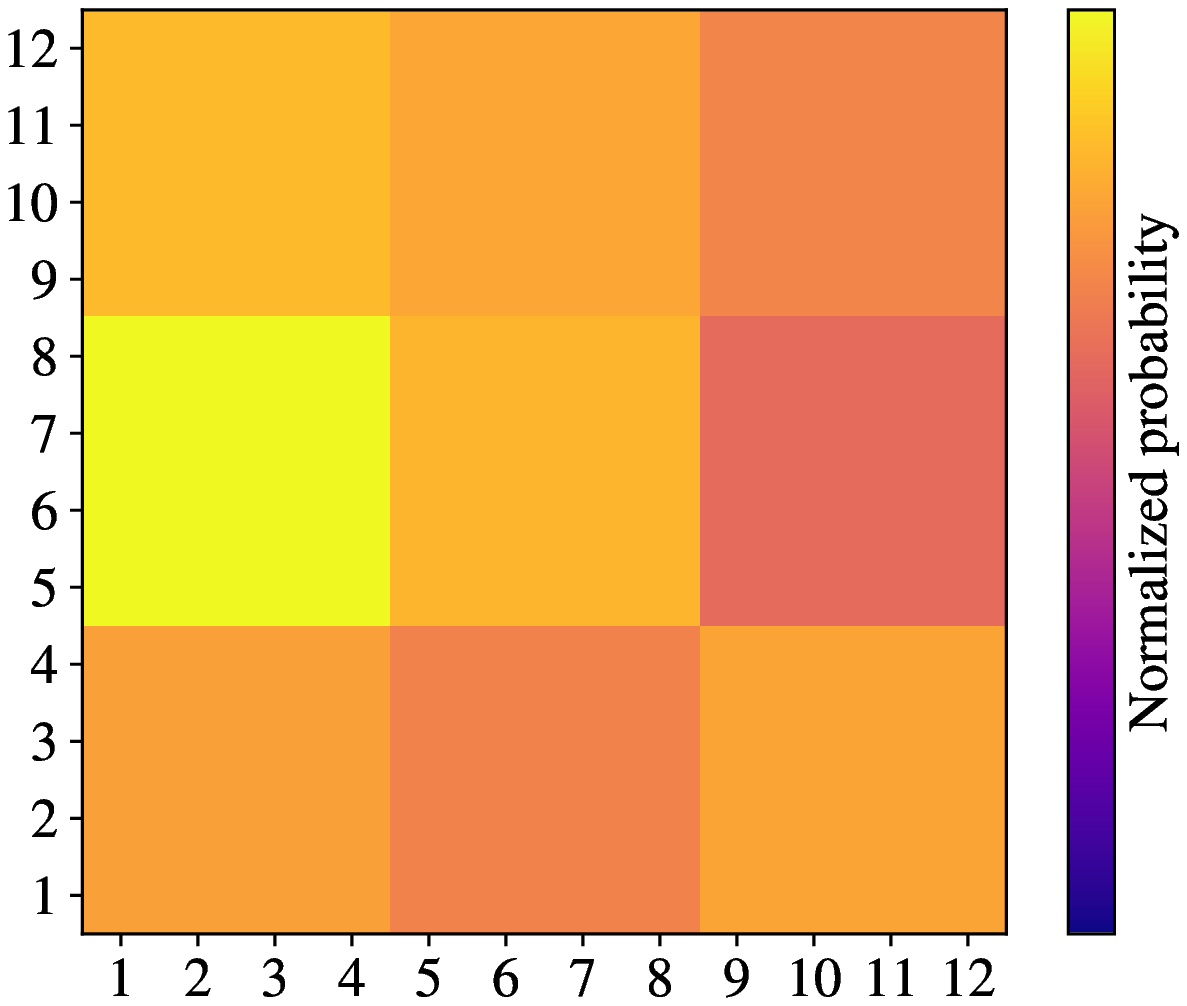}}
    \subfloat[%
  Sensor transmission (\gls{cjcc}). \label{fig:transmission_cjcc_1}%
]{\includegraphics[width=0.24\linewidth]{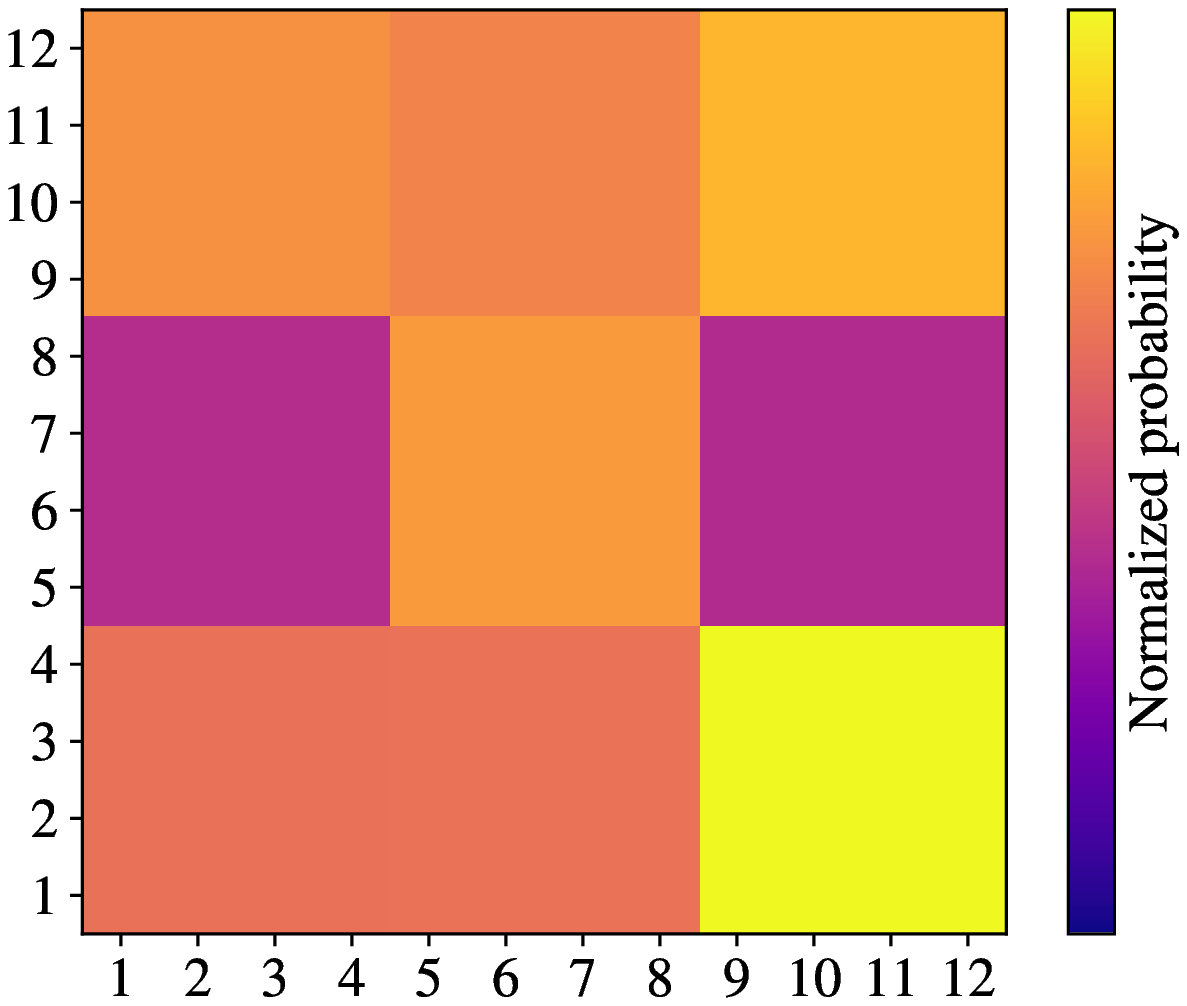}}
    \subfloat[%
  Sensor transmission (\gls{oc}). \label{fig:transmission_oc_1}%
  ]{\includegraphics[width=0.24\linewidth]{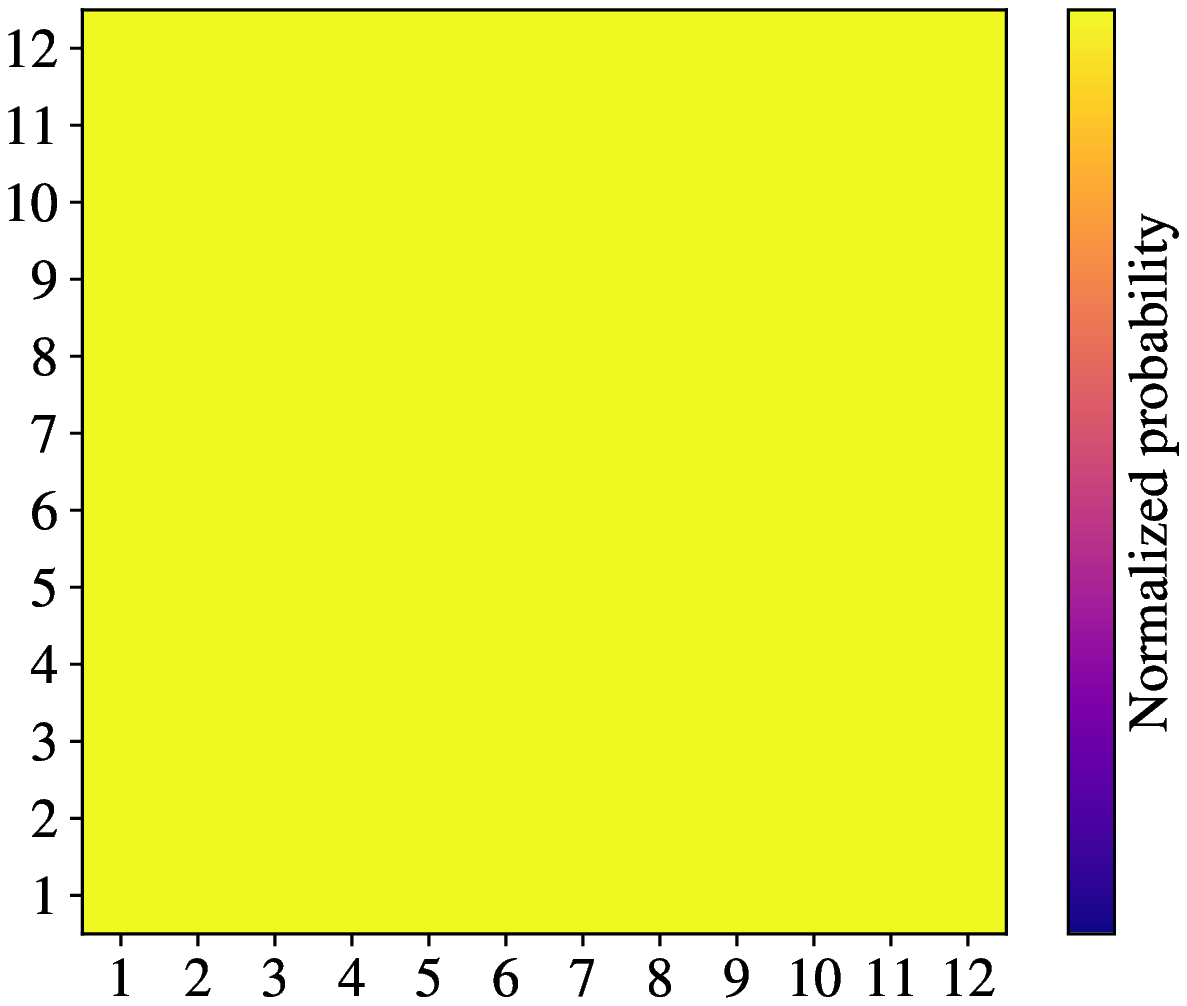}}
\caption{Heatmaps of the \gls{auv} location and transmitted area probability in the debris avoidance task.}
\label{fig:heatmap_1}
\end{figure*}

As expected, \gls{oc} outperforms all other strategies, with the lowest values of $N_{\text{step}}$.
In the obstacle avoidance scenario, the \gls{auv} accomplishes its goal in less than $30$ steps in half of the testing episodes, while the distribution median decreases to $22$ steps in the data muling case. This is because the \gls{oc} scheme allows the \gls{auv} to perfectly know the environment, making the problem trivial to solve.  
On the opposite extreme, the \gls{rc} strategy leads to the worst results, as the buoy transmissions do not take into consideration the relevance of the information to the \gls{auv} when choosing what to transmit. This leads to a median task completion time of $39$ steps in the data muling scenario, while the \gls{auv} fails more than 50\% of the test episodes in the obstacle avoidance scenario, not reaching the vessel even after 100 steps. Finally, the \gls{cc} scheme has an intermediate performance, taking 32 steps to finish the median episode in the data muling scenario and 57 in the obstacle avoidance scenario. 

However, the benchmark strategies are significantly outperformed by the \gls{jcc} schemes, which can almost get the same performance as the ideal communication setup. In particular, both \gls{djcc} and \gls{cjcc} can achieve superior performance in the obstacle avoidance mission, coordinating communications so as to always deliver the most relevant information to the \gls{auv}. On the other hand, the data muling scenario is more difficult, and even more so for a distributed setup: as buoys often only know the position of one of the two nodes, the risk of collisions is high, and the buoys will tend to behave more conservatively. In the latter scenario, the median value of $N_{\text{step}}$ is $26$ and $30$ for \gls{cjcc} and \gls{djcc}, respectively, which is still significantly lower than the benchmarks. This is even more noticeable when looking at the worst-case performance, represented by the upper whiskers of the boxplot: the difference between the schemes is extremely stark in this case, showing the benefits of joint control and communications and of having centralized knowledge of the state of the environment.

\begin{figure*}[t]
\centering
    \subfloat[%
  Robot location (\gls{cc}). \label{fig:location_cc_2}%
]{\includegraphics[width=0.24\linewidth]{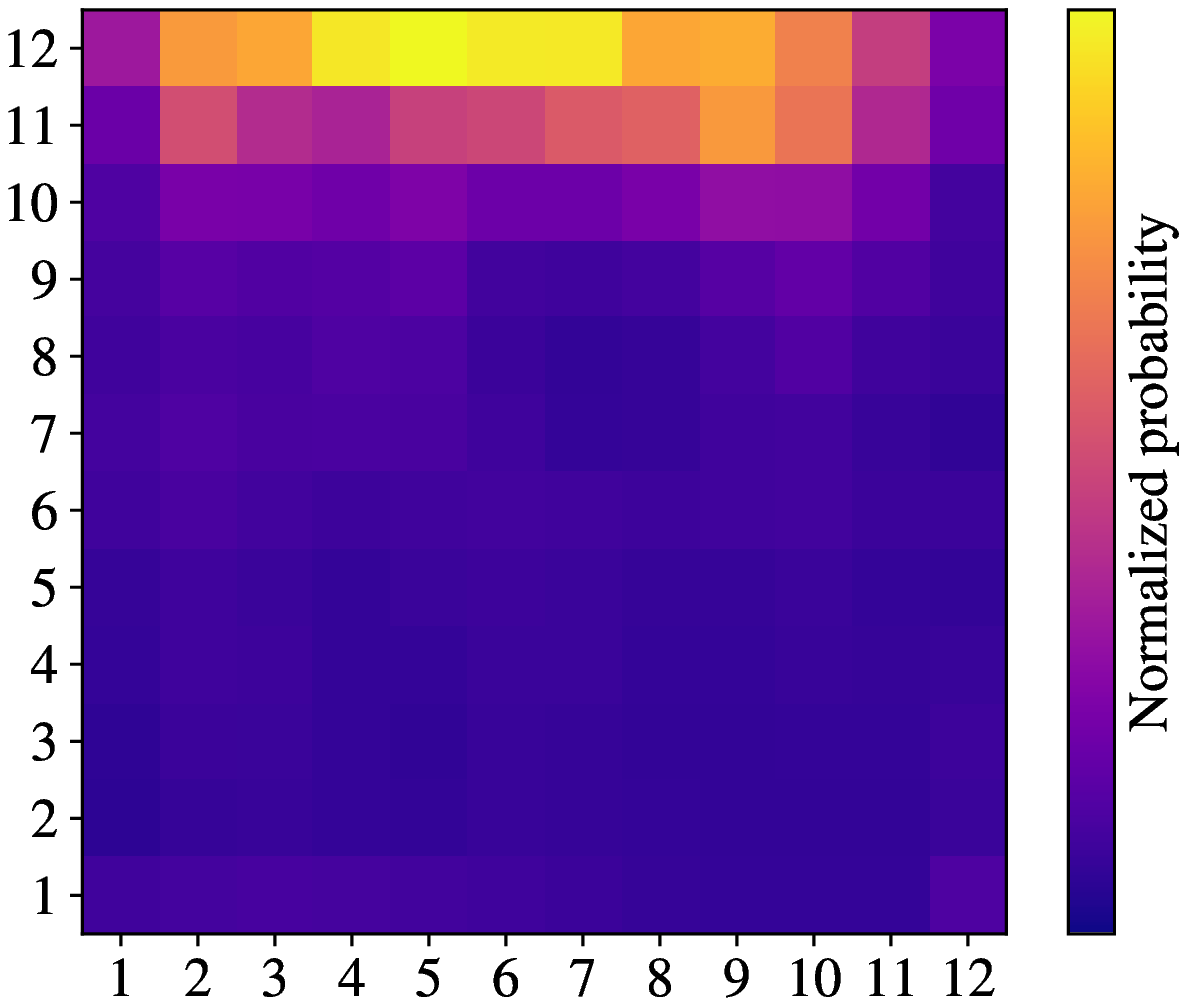}}
    \subfloat[%
  Robot location (\gls{djcc}). \label{fig:location_djcc_2}%
]{\includegraphics[width=0.24\linewidth]{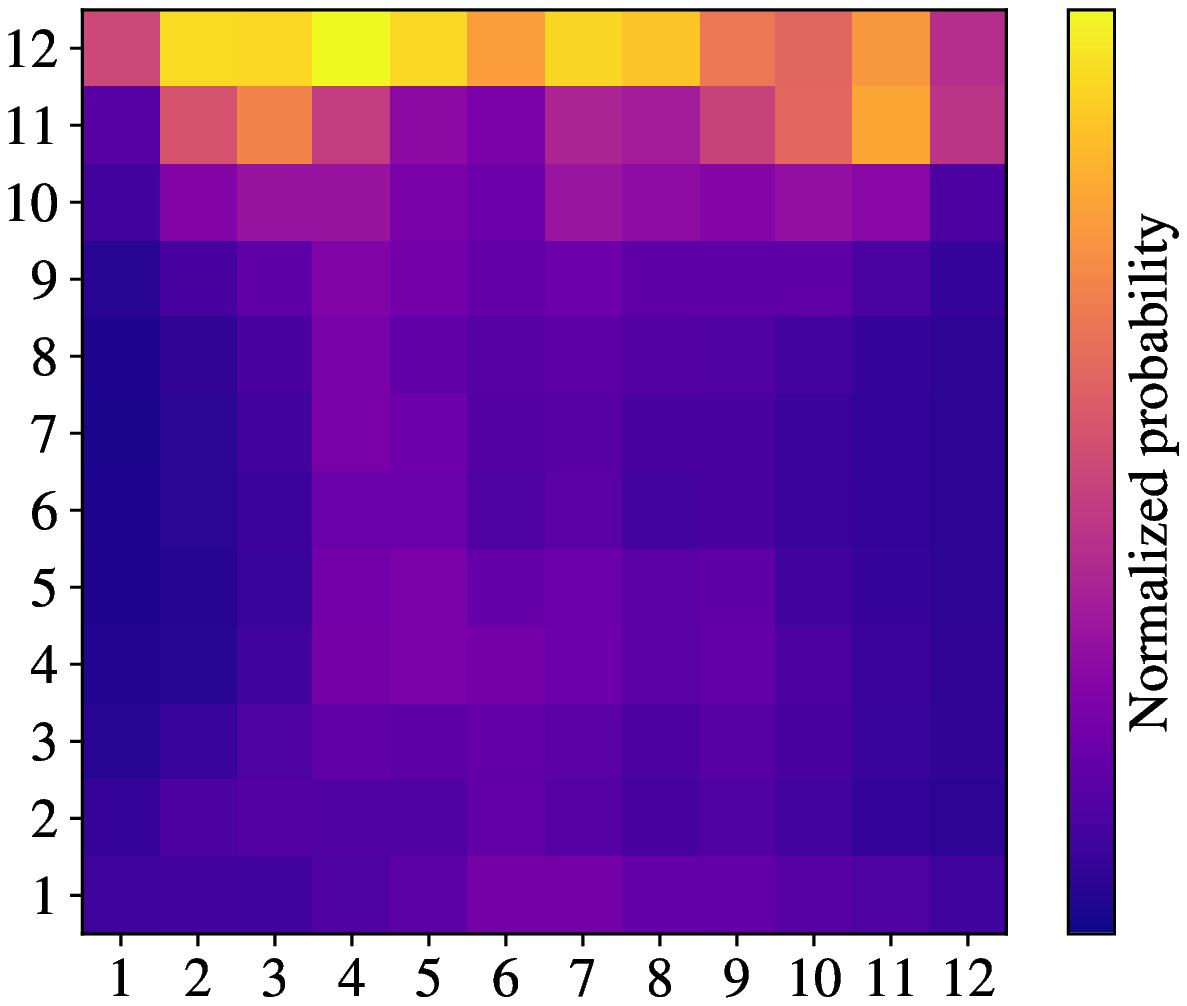}}
    \subfloat[%
  Robot location (\gls{cjcc}). \label{fig:location_cjcc_2}%
]{\includegraphics[width=0.24\linewidth]{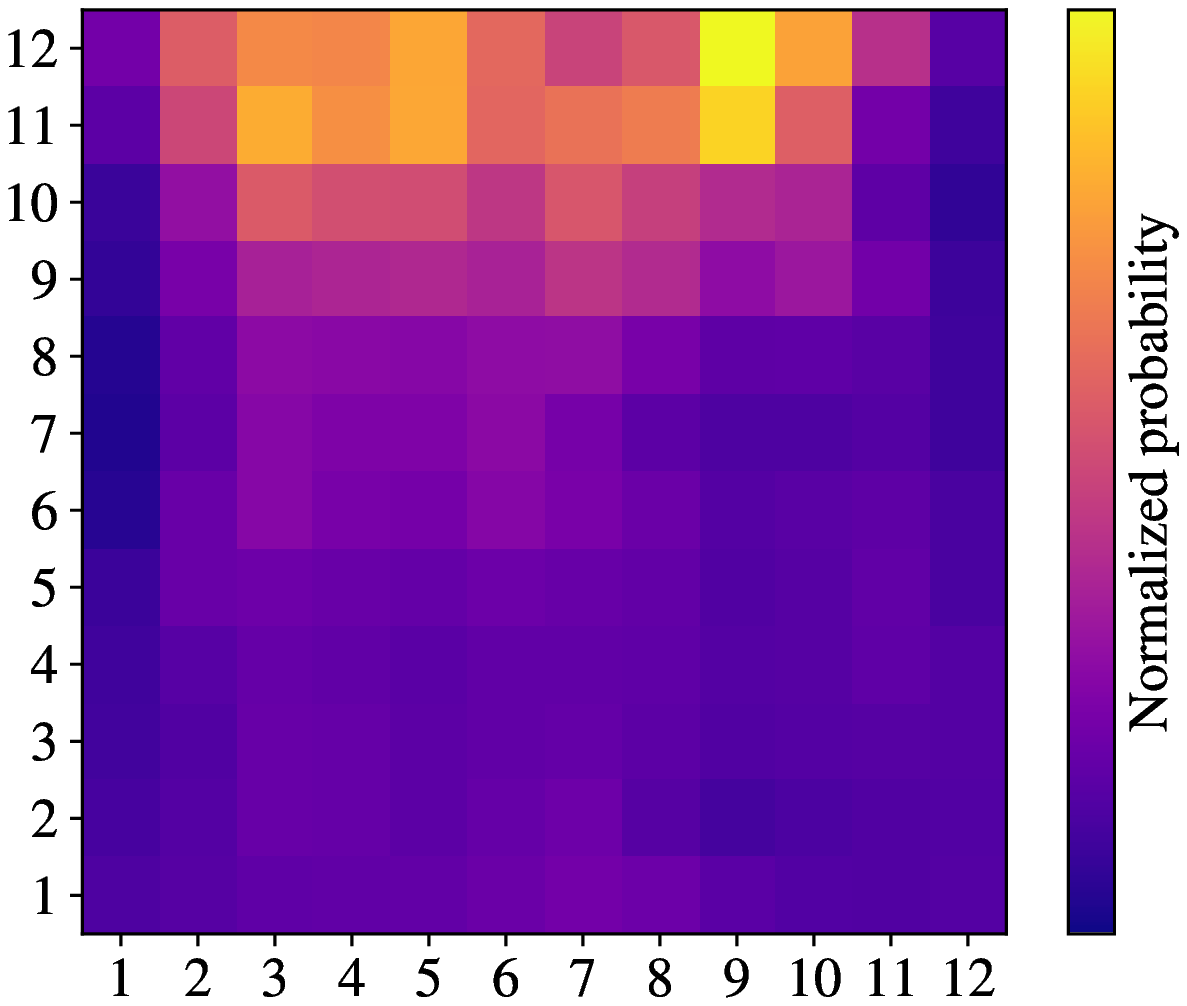}}
    \subfloat[%
  Robot location (\gls{oc}). \label{fig:location_oc_2}%
]{\includegraphics[width=0.24\linewidth]{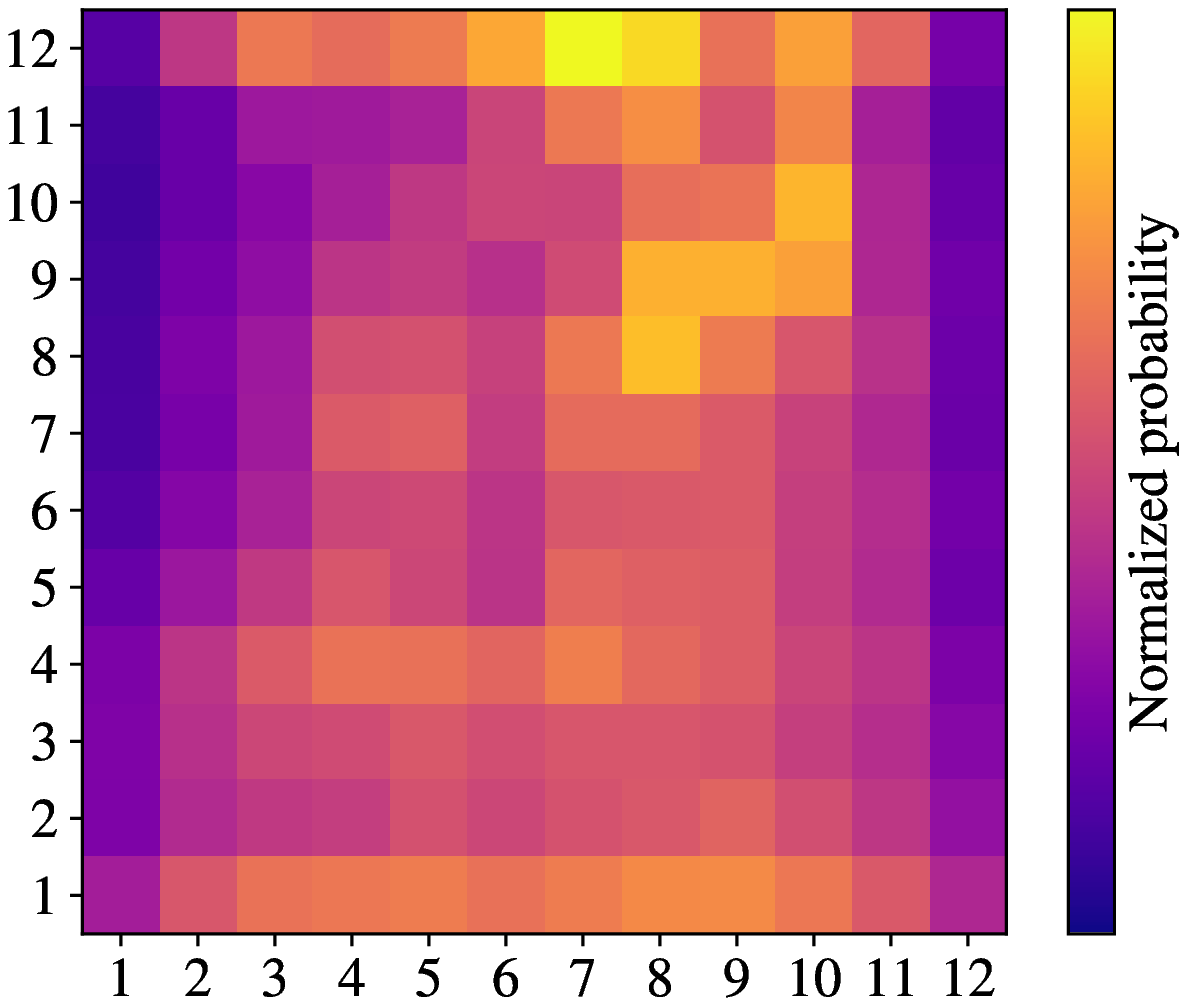}}\\
    \subfloat[%
  Sensor transmission (\gls{cc}). \label{fig:transmission_cc_2}%
]{\includegraphics[width=0.24\linewidth]{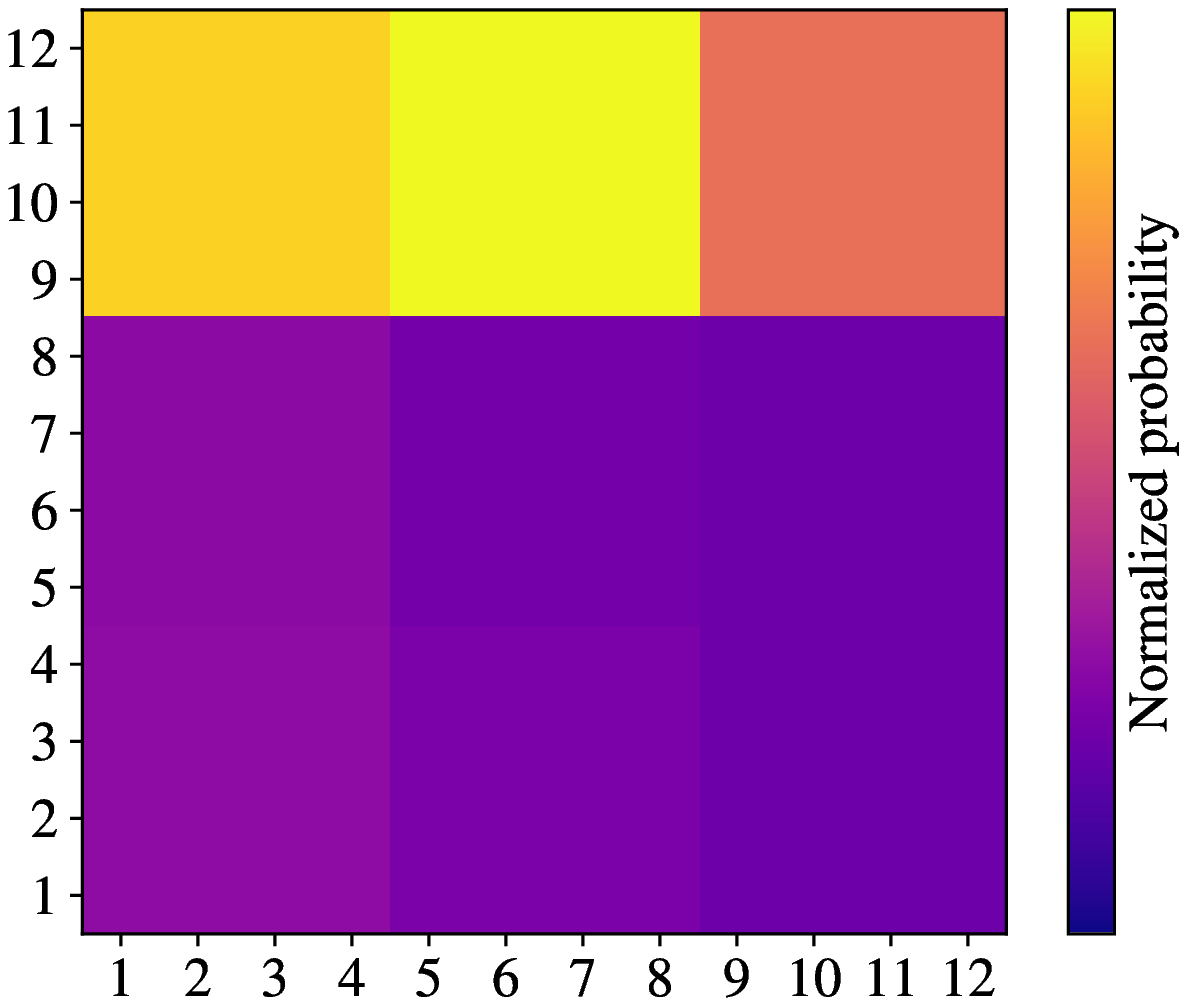}}
    \subfloat[%
  Sensor transmission (\gls{djcc}). \label{fig:transmission_djcc_2}%
]{\includegraphics[width=0.24\linewidth]{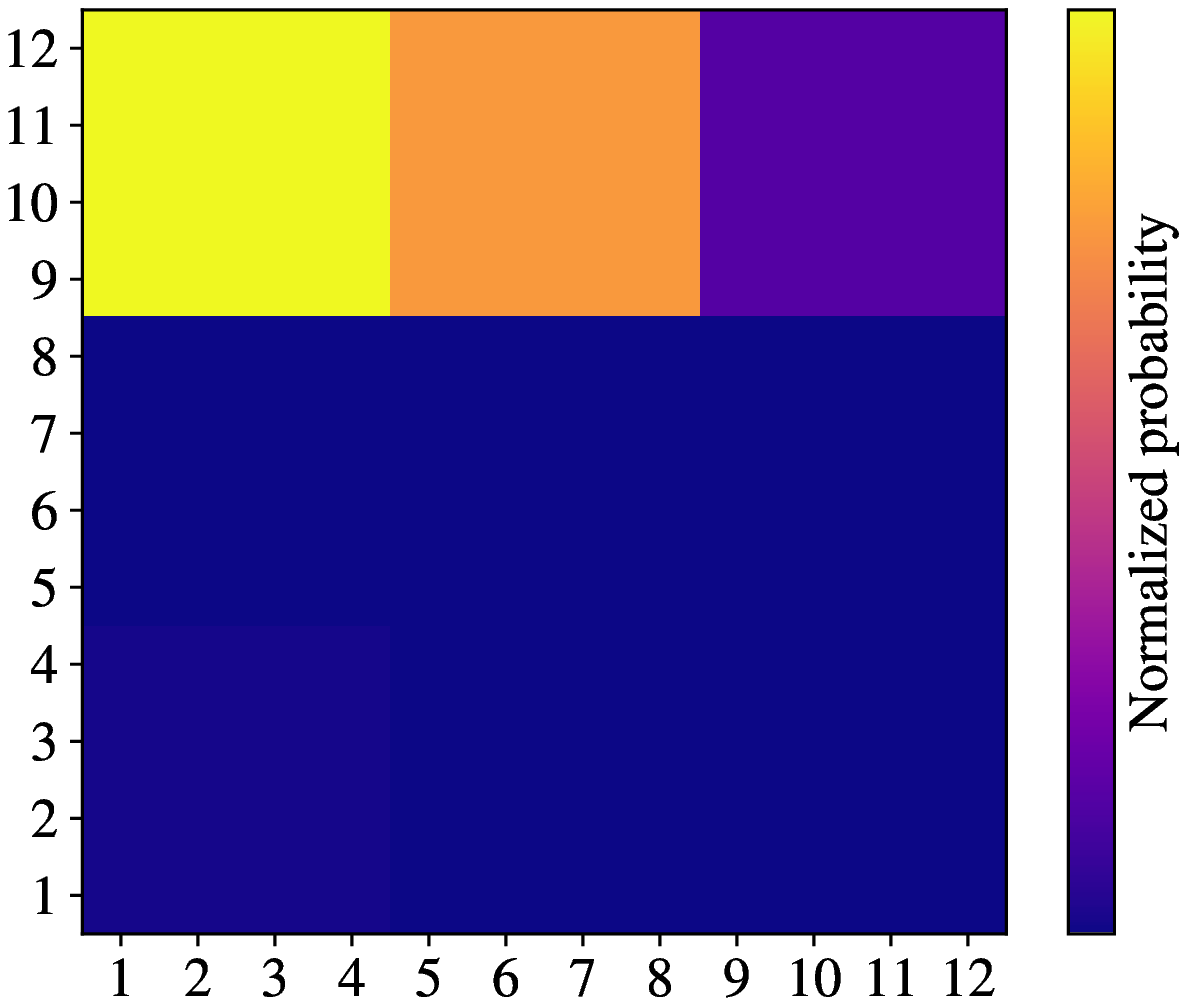}}
    \subfloat[%
  Sensor transmission (\gls{cjcc}). \label{fig:transmission_cjcc_2}%
]{\includegraphics[width=0.24\linewidth]{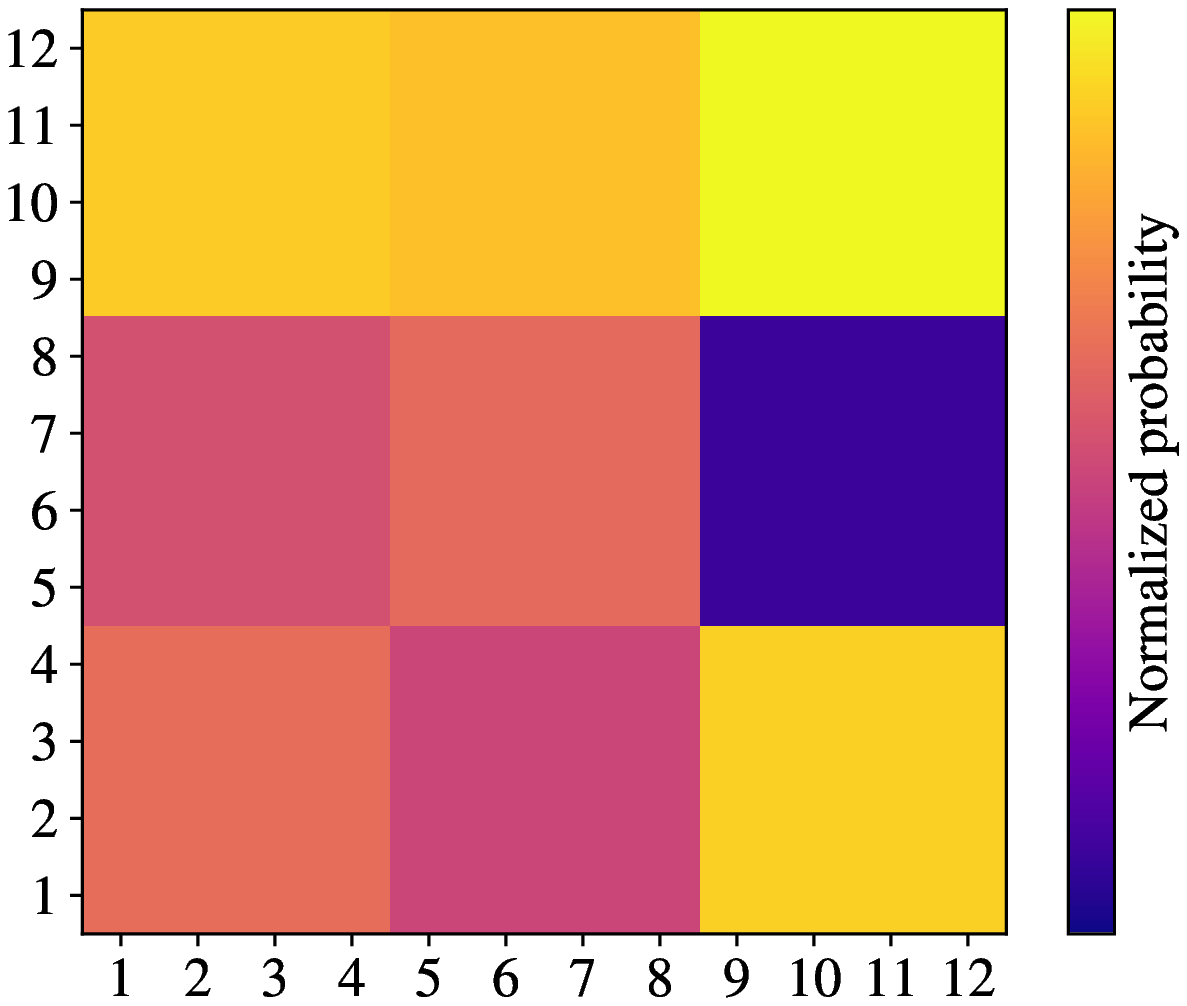}}
    \subfloat[%
  Sensor transmission (\gls{oc}). \label{fig:transmission_oc_2}%
  ]{\includegraphics[width=0.24\linewidth]{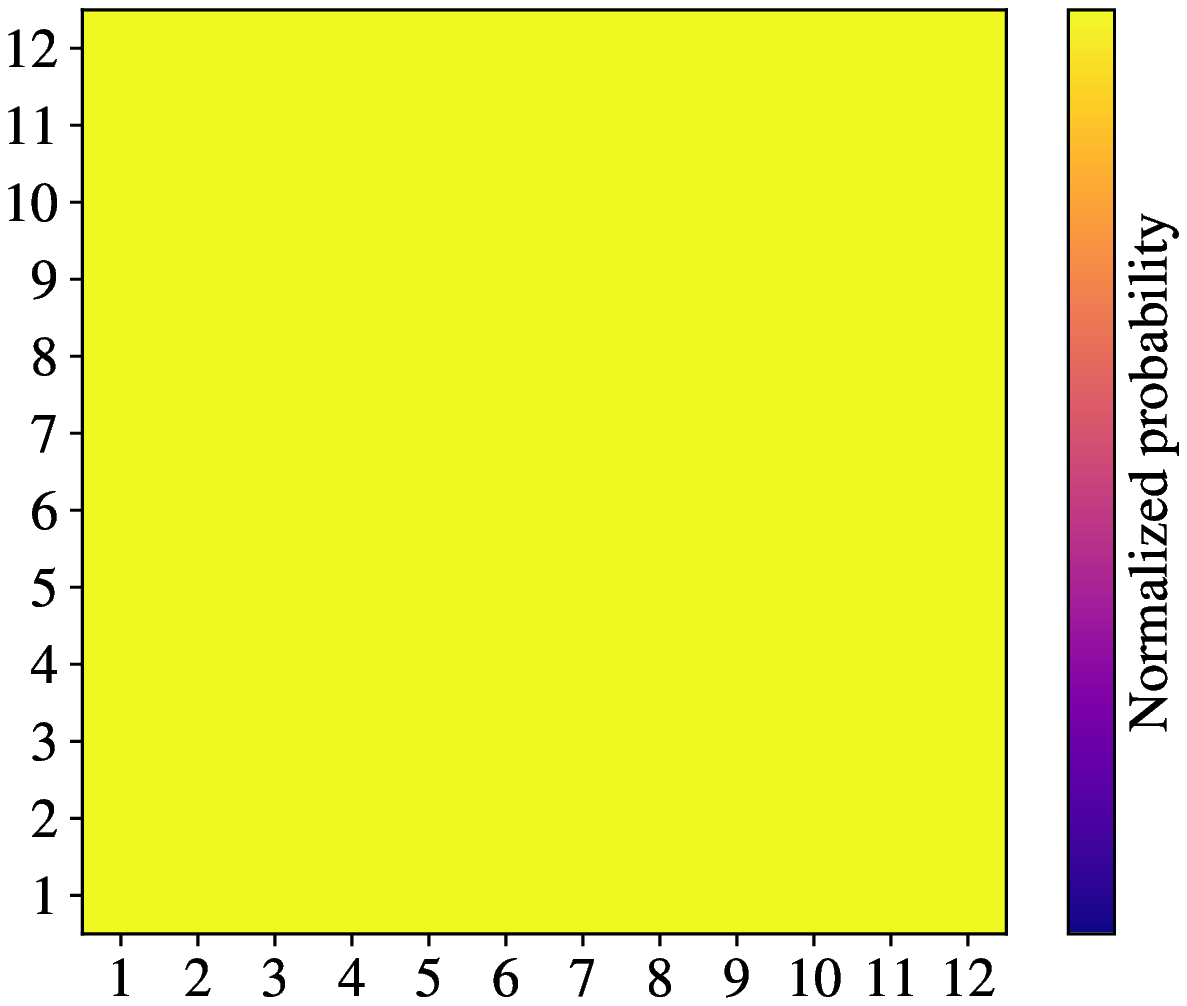}}
\caption{Heatmaps of the \gls{auv} location and transmitted area probability in the data muling task.}
\label{fig:heatmap_2}
\end{figure*}

To better illustrate the agents' behavior, Figs.~\ref{fig:location_cc_1}-\subref*{fig:location_oc_1} show the distribution of the \gls{auv} position in the obstacle avoidance scenario with different control strategies. The distribution is presented as a colormap: the lighter the cell color, the higher the probability of the \gls{auv} to be in the corresponding position during the testing phase. We can notice that the \gls{auv} tends to keep close to the center of the map when using the \gls{cc} scheme, never straying too far to the left or right: on the other hand, the \gls{jcc} strategies have a distribution more similar to the \gls{oc} case. This is due to the influence of transmissions from the buoys: if the \gls{auv} can know where the opening is, it can move directly toward it, instead of staying close to the center of the map and then gradually exploring. While this is a relatively simple example, and constructing a better transmission heuristic might be possible, the joint design can optimize both strategies jointly without any additional design requirements. 

We can see the same pattern in Figs.~\ref{fig:transmission_cc_1}-\subref*{fig:transmission_oc_1}, which shows the transmission patterns for the buoys: while the buoys closer to the center of the map transmit much more often when using the \gls{cc} scheme, buoys that actually have relevant information are much more active in the two \gls{jcc} schemes, helping the \gls{auv} reach the target faster.

Finally, Fig.~\ref{fig:heatmap_2} shows the distribution of the \gls{auv} positions and the buoys' transmissions in the data muling scenario. As for the previous case, we can appreciate that the \gls{auv} takes more complex trajectories in the case of the \gls{jcc} approaches, as shown in Figs.~\ref{fig:location_djcc_2}--\ref{fig:location_cjcc_2}, than when using the \gls{cc} strategy, as shown in Fig.\ref{fig:location_cc_2}. In this case, there is still a gap between the \gls{jcc} strategies and the \gls{oc} scheme, as coordination is much harder, and we can see that the trajectories for \gls{cjcc} are much more evenly distributed than for \gls{djcc}. This is explained by the transmission distribution in Figs.~\ref{fig:transmission_djcc_2}--\ref{fig:transmission_cjcc_2}: while centralized communication can always deliver relevant information, the buoys in the \gls{djcc} scenario cannot coordinate effectively when the \gls{auv} is in the lower portion of the map, i.e., at the beginning of the task, when neither node has been visited. In order to avoid collisions, which are heavily penalized by the reward function, the buoys tend to only transmit when the \gls{auv} is already close to the target, helping it reach the second node when necessary. This also explains the performance gap between \gls{cjcc} and \gls{djcc}, as both the transmission and location distribution maps for \gls{djcc} are in between \gls{cc} and \gls{cjcc}.

\section{Conclusions and future work}\label{sec:concl}

In this paper, we developed a new theoretical framework to model the communication among multiple agents operating in the same learning environment.
Our system, named \gls{cppomdp}, includes two classes of agents: the robots, which have to perform a physical task, and the sensors, which can sense the environment and transmit information to the robots.
Since their reward depends directly on the robot's actions, the sensors are encouraged to identify the best communication policy to help the robots complete their task as effectively as possible. The model is extremely flexible, and can represent a plethora of practical scenarios, while still limiting complexity. We tested the described framework in a reference scenario, in which an \gls{auv} needs to complete a mission in an underwater environment, while a set of buoys can transmit information about other areas of the map, widening its perception. Our simulations show that the joint training of the buoys and the \gls{auv}, based on our \gls{cppomdp} model, significantly outperforms classical networked control strategies.

Future work on the subject might involve more complex scenarios with multiple robots and sensors, in which information is distributed across the agents, none of which has a full view of the scenario. This leads to interesting connections to open machine learning problems, as agents have to develop a theory of mind to estimate not just the state, but also the knowledge of other agents, and to game theoretical issues, as the behavior of the swarm of robots as a whole is determined by game theoretical principles. The case in which agents have partially conflicting objectives is particularly interesting, as sensors might find it beneficial to transmit false or misleading information, introducing questions of trust and reciprocity to the problem.

\bibliographystyle{IEEEtran}
\bibliography{biblio.bib}

\appendix

\subsection*{Debris Avoidance Scenario Definition}\label{app:debris}

In the debris avoidance scenario, the obstacles are arranged in three horizontal lines on the map, as shown in Fig.~\ref{fig:debris}. Each line of debris has a single opening, whose location can change over time. We denote the coordinates of the free passages on the $x$-axis at slot $k$ by $z_{0}(k)$, $z_{1}(k)$, and $z_{2}(k)$, respectively. Hence, at the beginning of each slot $k$, the set of positions containing obstacles is given by:
\begin{equation}
\begin{aligned}
 M_k(\mathbf{x})=\omega\iff x^{(1)}(k)\neq z^{(1)}_j(k),x^{(2)}(k)\in\mathcal{Y}_j,
\end{aligned}
\end{equation}
where $\mathcal{Y}_j$ is the set containing the coordinates of the $j$-th debris line on the $y$ axis:
\begin{equation}
  \begin{aligned}
    \mathcal{Y}_0=\{2,3\},\mathcal{Y}_1=\left\{\frac{N}{2},\frac{N}{2}+1\right\},\mathcal{Y}_2=\{N-2,N-1\}.
  \end{aligned}
\end{equation}
At the beginning of each episode, the free passages' coordinates are generated from a discrete uniform distribution in $\mathcal{N}$.
The locations $z_{0}(k)$, $z_{1}(k)$, and $z_{2}(k)$ are then updated following the same periodicity as the buoys' communication.
Hence, at the beginning of any slot $k$ such that $\text{mod}(k,K_p) = 0$, the new values of $z_{0}(k)$, $z_{1}(k)$, and $z_{2}(k)$ are picked from the distribution
\begin{equation}
    P(z_{j}(k)=z) = \frac{(N - |z - z_{j}(k-1)|)^2}{\sum_{q=1}^N (N - |q - z_{j}(k-1)|)^2}.
\end{equation}

In this scenario, the intermediate reward indicator function $\chi(s(k),a_{\ell}(k))$ is tuned to give the robot a small reward if it makes progress towards the vessel;

needs to be tuned to encourage the \gls{auv} to avoid the debris and finally reach the vessel.
In particular, the \gls{auv} receives a reward $\rho \in\mathbb{R}^+$ when it reaches the surface vessel, i.e., if $\mathbf{x}_{\ell,k+1}=\mathbf{x}_{v}$.
We also assign a smaller, positive intermediate reward $\sigma < \rho$ if the \gls{auv} makes progress its task, i.e., in one of these cases:
\begin{enumerate}
    \item If $\mathbf{a}_{\ell}(k)=(0,1)$ and $M_{k}(\mathbf{x}_{\ell}(k) + \mathbf{a}_{\ell}(k)) = \phi$, i.e., if the \gls{auv} moves upward in the map, without entering in the area occupied by the debris;
    \item If $|\mathbf{x}_{\ell}^{(1)}(k) + \mathbf{a}_{\ell}^{(1)}(k) - z_{j}(k)| < |\mathbf{x}_{\ell}^{(1)}(k) - z_{j}(k)|$ and $\mathbf{x}_{\ell}^{(2)}(k)+1\in\mathcal{Y}_j$, i.e., if the \gls{auv} has reached a debris line, and moves horizontally towards a free passage;
    \item If $||\mathbf{x}_{\ell}(k) + \mathbf{a}_{\ell}(k) - \mathbf{x}_{f}||_2 < ||\mathbf{x}_{\ell}(k) - \mathbf{x}_{f}||_2$ and $x_{\ell}^{(2)}(k)=N$, i.e., if the \gls{auv} is on the top row of the map, and moves horizontally towards the vessel.
\end{enumerate}
We can then summarize these conditions in the Boolean function $\Phi(\mathbf{x}_{\ell, k}$, $\mathbf{x}_{v}$, $\mathbf{a}_{\ell, k}$, $\mathcal{M}_{k})$, which returns a true value if one of the above conditions is verified.

\subsection*{Data Muling Scenario Definition}\label{app:muling}

In the data muling scenario, we define an \emph{active target} as any target node that has not been visited by the \gls{auv} yet. We will only consider active targets, as the ones that have already been visited do not affect the reward and are removed from $\mathbf{M}_k$. The set of active targets in slot $k$ is denoted by $\mathcal{T}_k$.

At the beginning of each episode, the targets' coordinates are generated from a discrete uniform distribution in $\mathcal{M}$.
The position of the $i$-th target $\mathbf{t}_{i}(k)$ then changes following the same periodicity as the buoys' communications. Whenever $\text{mod}(k,K_p) = 0$, the target's position is updated according to the following distribution:
\begin{equation}
    P(\mathbf{t}_{i}(k+1)=\mathbf{x}) =\prod_{i=1}^2 \frac{(N - |x^{(j)} - t^{(j)}_{i}(k)|)^2}{\sum_{n \in \mathcal{N}} (N - |n - x^{(j)}|)^2},
\end{equation}
where $x^{(j)}$ is the $j$-th coordinate of $\mathbf{x}$.

In this scenario, the intermediate reward indicator function $\chi(s(k),a_{\ell}(k))$ is designed to make the \gls{auv} visit all targets, then reach the vessel in the shortest possible time. The function is equal to 1 if any of the following conditions are true:
\begin{enumerate}
    \item If $|\mathcal{T}_{k}|=2$, $||\mathbf{x}_{\ell}(k) + \mathbf{a}_{\ell}(k) - \mathbf{t}_{i, k}||_2 < ||\mathbf{x}_{\ell}(k) - \mathbf{t}_{i, k}||_2$, and $||\mathbf{x}_{\ell}(k) - \mathbf{t}_{3-i, k}||_2 + ||\mathbf{t}_{3-i, k} - \mathbf{x}_{v}||_2 < ||\mathbf{x}_{\ell}(k) - \mathbf{t}_{3-i, k}||_2 + ||\mathbf{t}_{3-i, k} - \mathbf{x}_{v}||_2$, with $i\in\{1,2\}$, i.e., if both targets are still unvisited, the \gls{auv} is moving towards $\mathbf{t}_{i}(k)$, and it is more convenient to visit the node in $\mathbf{t}_{i}(k)$ to end the mission in the shortest time;
    \item If $|\mathcal{T}_k|=1$ and $|\mathbf{x}_{\ell}(k) + \mathbf{a}_{\ell}(k) - \mathbf{t}_{1}(k)||_2 < ||\mathbf{x}_{\ell}(k) - \mathbf{t}_{1}(k)||_2$, i.e., if only a single node is still unvisited and the \gls{auv} is moving towards its position;
    \item If $|\mathcal{T}_{k}|=0$ and $|\mathbf{x}_{\ell}(k) + \mathbf{a}_{\ell}(k) - \mathbf{x}_{f}| < |\mathbf{x}_{\ell}(k) - \mathbf{x}_{f}|$, i.e., if all the targets have been visited and the \gls{auv} is moving towards the vessel.
\end{enumerate}

\end{document}